\documentclass[10pt,twocolumn,letterpaper]{article}

\usepackage{cvpr}              

\usepackage{graphicx}
\usepackage{amsmath}
\usepackage{amssymb}
\usepackage{booktabs}
\usepackage{blindtext}
\usepackage{caption}

\usepackage{multirow}
\usepackage{bbm}
\usepackage{bm}

\usepackage[dvipsnames]{xcolor}

\definecolor{cvprblue}{rgb}{0.21,0.49,0.74}
\usepackage[pagebackref,breaklinks,colorlinks,citecolor=cvprblue]{hyperref}

\def\paperID{10650} 
\def\confName{CVPR}
\def\confYear{2024}

\title{DiffusionGAN3D: Boosting Text-guided 3D Generation and Domain Adaptation by Combining 3D GANs and Diffusion Priors}

\author{Biwen Lei, Kai Yu, Mengyang Feng, Miaomiao Cui, Xuansong Xie\\
Alibaba Group\\
{\tt\small \{biwen.lbw, jinmao.yk, mengyang.fmy, miaomiao.cmm\}@alibaba-inc.com, }\\
{\tt\small xingtong.xxs@taobao.com}
}
\begin{document}


\twocolumn[{
\renewcommand\twocolumn[1][]{#1}
\maketitle
\vspace{-30pt}
\begin{center}
    \centering
    \captionsetup{type=figure}
    \includegraphics[scale=0.47]{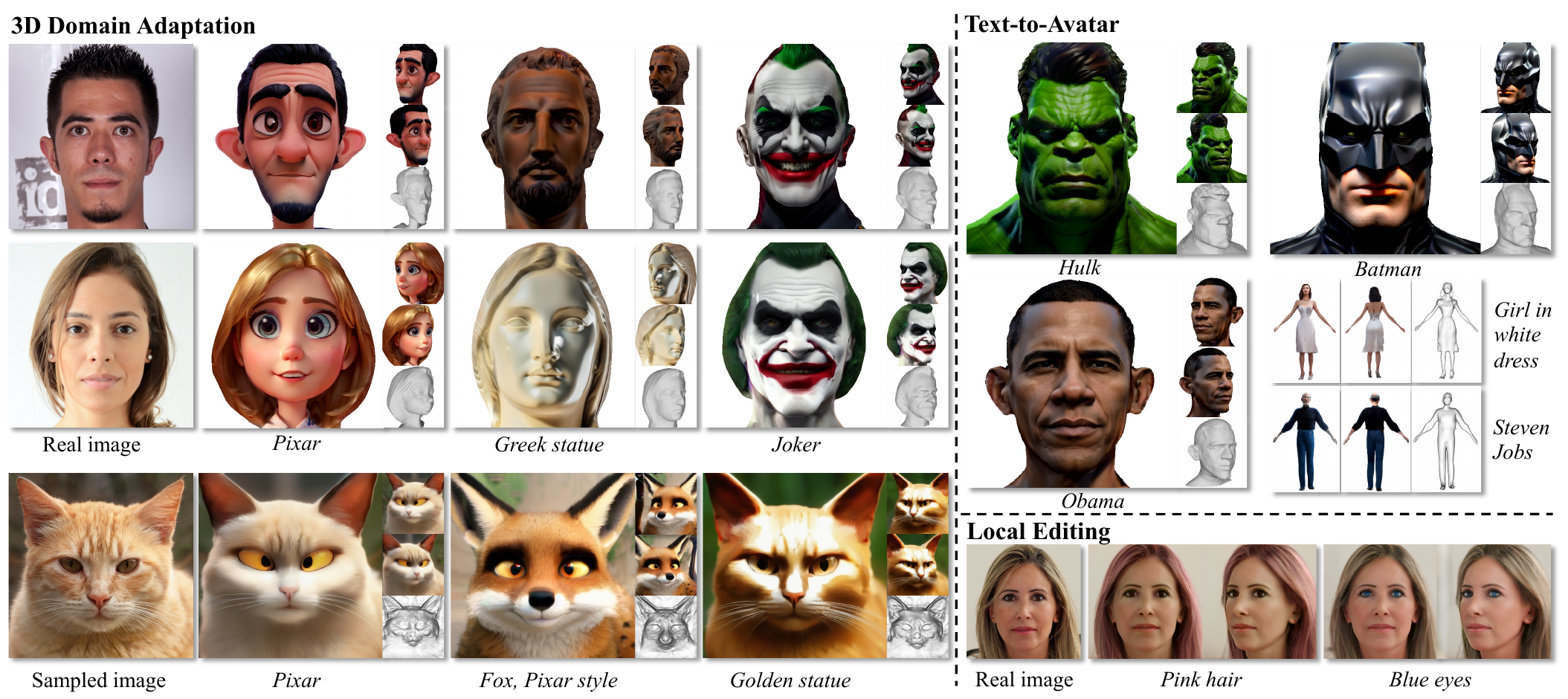}
    \vspace{-10pt}
    \captionof{figure}{Some results of the proposed DiffusionGAN3D on different tasks.}
    \label{fig: teaser}
\end{center}
}]


\begin{abstract}
\vspace{-13pt}

Text-guided domain adaptation and generation of 3D-aware portraits find many applications in various fields. However, due to the lack of training data and the challenges in handling the high variety of geometry and appearance, the existing methods for these tasks suffer from issues like inflexibility, instability, and low fidelity. In this paper, we propose a novel framework DiffusionGAN3D, which boosts text-guided 3D domain adaptation and generation by combining 3D GANs and diffusion priors. Specifically, we integrate the pre-trained 3D generative models (e.g., EG3D) and text-to-image diffusion models. The former provides a strong foundation for stable and high-quality avatar generation from text. And the diffusion models in turn offer powerful priors and guide the 3D generator finetuning with informative direction to achieve flexible and efficient text-guided domain adaptation. To enhance the diversity in domain adaptation and the generation capability in text-to-avatar, we introduce the relative distance loss and case-specific learnable triplane respectively. Besides, we design a progressive texture refinement module to improve the texture quality for both tasks above. Extensive experiments demonstrate that the proposed framework achieves excellent results in both domain adaptation and text-to-avatar tasks, outperforming existing methods in terms of generation quality and efficiency. The project homepage is at \href{https://younglbw.github.io/DiffusionGAN3D-homepage/}{https://younglbw.github.io/DiffusionGAN3D-homepage/}.
\end{abstract}
\vspace{-17pt}

\section{Introduction}
\label{sec:intro}

3D portrait generation and stylization find a vast range of applications in many scenarios, such as games, advertisements, and film production. While extensive works \cite{chan2022efficient, an2023panohead, dong2023ag3d, hong2022eva3d} yield impressive results on realistic portrait generation, the performance on generating stylized, artistic, and text-guided 3D avatars is still unsatisfying due to the lack of 3D training data and the difficulties in modeling highly variable geometry and texture.

Some works \cite{zhou2021cips, abdal20233davatargan, zhang2023styleavatar3d, zhang2023deformtoon3d, kim2023datid, kim2023podia, song2023agilegan3d} perform transfer learning on a pre-trained 3D GAN generator to achieve 3D stylization, which relies on a large number of stylized images and strictly aligned camera poses for training. \cite{abdal20233davatargan, song2023agilegan3d} leverage existing 2D-GAN trained on a specific domain to synthesize training data and implement finetuning with adversarial loss. In contrast, \cite{zhang2023styleavatar3d, kim2023datid, kim2023podia} utilize text-to-image diffusion models to generate training datasets in the target domain. This enables more flexible style transferring but also brings problems like pose bias, tedious data processing, and heavy computation costs. Unlike these adversarial finetuning based methods, StyleGAN-Fusion \cite{song2022diffusion} adopts SDS \cite{poole2022dreamfusion} loss as guidance of text-guided adaptation of 2D and 3D generators, which gives a simple yet effective way to fulfill domain adaptation. However, it also suffers from limited diversity and suboptimal text-image correspondence.

The recently proposed Score Distillation Sampling (SDS) algorithm \cite{poole2022dreamfusion} exhibits impressive performance in text-guided 3D generation. Introducing diffusion priors into the texture and geometry modeling notably reduces the training cost and offers powerful 3D generation ability. However, it also leads to issues like unrealistic appearance and Janus (multi-face) problems. Following \cite{poole2022dreamfusion}, massive works \cite{DreamAvatar, avatarcraft, dreamhuman, Avatarverse, lin2023magic3d, wang2023vsd, tang2023dreamgaussian} have been proposed to enhance the 
generation quality and stability. Nevertheless, the robustness and visual quality of the generated model are still far less than the current generated 2D images.

\begin{figure*}[t]
   \setlength{\belowcaptionskip}{-0.3cm}
   \setlength{\abovecaptionskip}{0.1cm}
\centerline{\includegraphics[scale=0.423]{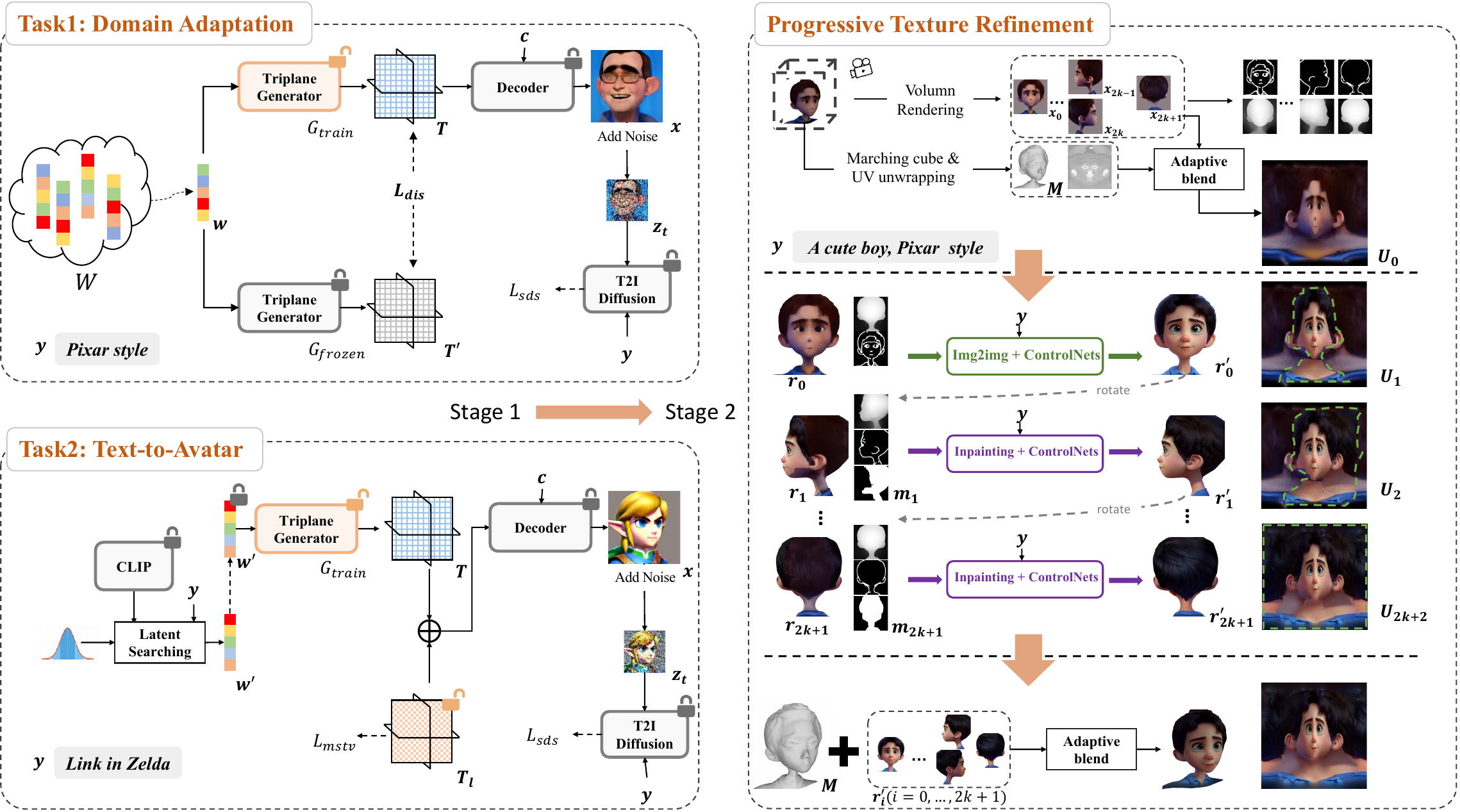}}
\caption{Overview of the proposed two-stage framework DiffusionGAN3D.}
\vspace{-6pt}
\label{fig:framework}
\end{figure*}

Based on the observations above, we propose a novel two-stage framework DiffusionGAN3D to boost the performance of 3D domain adaptation and text-to-avatar tasks by combining 3D generative models and diffusion priors, as shown in Fig.~\ref{fig:framework}. For the text-guided 3D \textbf{Domain Adaptation} task, we first leverage 
diffusion models and adopt 
SDS loss to finetune a pre-trained EG3D-based model \cite{chan2022efficient, an2023panohead, dong2023ag3d} with random noise input and camera views. The relative distance loss is introduced to deal with the loss of diversity caused by the SDS technique. Additionally, we design a diffusion-guided reconstruction loss to adapt the framework to local editing scenarios. Then, we extend the framework to \textbf{Text-to-Avatar} task by finetuning 3D GANs with a fixed latent code that is obtained guided by CLIP \cite{radford2021learning} model. During optimization, a case-specific learnable triplane is introduced to strengthen the generation capability of the network. To sum up, in our framework, the diffusion models offer powerful text-image priors, which guide the domain adaptation of the 3D generator with informative direction in a flexible and efficient way. In turn, 3D GANs provide a strong foundation for text-to-avatar, enabling stable and high-quality avatar generation. Last but not least, taking advantage of the powerful 2D synthesis capability of diffusion models, we propose a \textbf{Progressive Texture Refinement} module as the second stage for these two tasks above, which significantly enhances the texture quality. Extensive experiments demonstrate that our method exhibits excellent performance in terms of generation quality and stability on 3D domain adaptation and text-to-avtar tasks, as shown in Fig.~\ref{fig: teaser}.

Our main contributions are as follows: \\
\textbf{(A)} We achieve text-guided 3D domain adaptation in high quality and diversity by combining 3D GANs and diffusion priors with the assistance of the relative distance loss. \\
\textbf{(B)} We adapt the framework to a local editing scenario by designing a diffusion-guided reconstruction loss. \\
\textbf{(C)} We achieve high-quality text-to-avatar in superior performance and stability by introducing the case-specific learnable triplane. \\
\textbf{(D)} We propose a novel progressive texture refinement stage, which fully exploits the image generation capabilities of the diffusion models and greatly enhances the quality of texture generated above.

\section{Related Work}


\noindent {\bf Domain Adaptation of 3D GANs.} The advancements in 3D generative models \cite{chan2022efficient, or2022stylesdf, an2023panohead, dong2023ag3d, chan2021pi, gadelha20173d, henzler2019escaping, gu2021stylenerf, liao2020towards, schwarz2020graf} have enabled geometry-aware and pose-controlled image generation. Especially, EG3D \cite{chan2022efficient} utilizes triplane as 3D representation and integrates StyleGAN2 \cite{karras2020analyzing} generator with neural rendering \cite{mildenhall2021nerf} to achieve high-quality 3D shapes and view-consistency image synthesis, which facilitates the downstream applications such as 3D stylization, GAN inversion \cite{lan2023self}. Several works \cite{jin2022dr, zhou2021cips, abdal20233davatargan, zhang2023deformtoon3d} achieve 3D domain adaptation by utilizing stylized 2D generator to synthesize training images or distilling knowledge from it. 
In contrast, \cite{kim2023datid, kim2023podia, zhang2023styleavatar3d} leverage the powerful diffusion models to generate training datasets in the target domain and accomplish text-guided 3D domain adaptation with great performance. Though achieving impressive results, these adversarial learning based methods above suffer from issues such as pose bias, tedious data processing, and heavy computation cost. Recently, non-adversarial finetuining methods \cite{alanov2022hyperdomainnet, gal2021stylegan, song2022diffusion} also exhibit great promise in text-guided domain adaptation. Especially, StyleGAN-Fusion \cite{song2022diffusion} adopts SDS loss as guidance for the adaptation of 2D generators and 3D generators. It achieves efficient and flexible text-guided domain adaptation but also faces the problems of limited diversity and suboptimal text-image correspondence.


\noindent {\bf Text-to-3D Generation.} In recent years, text-guided 2D image synthesis \cite{rombach2022high, Saharia2022PhotorealisticTD, zhang2023adding, ramesh2022, ruiz2023dreambooth} achieve significant progress and provide a foundation for 3D generation. Prior works, including CLIP-forge \cite{sanghi2022clip}, CLIP-Mesh \cite{2022clipmesh}, and DreamFields \cite{2022dreamfields}, employ CLIP \cite{radford2021learning} as guidance to optimize 3D representations such as meshes and NeRF \cite{mildenhall2021nerf}. DreamFusion \cite{poole2022dreamfusion} first proposes score distillation sampling (SDS) loss to utilize a pre-trained text-to-image diffusion model to guide the training of NeRF. It is a pioneering work and exhibits great promise in text-to-3d generation, but also suffers from over-saturation, over-smoothing, and Janus (multi-face) problem. Subsequently, extensive improvements \cite{lin2023magic3d, dreambooth3d, wang2023vsd, tang2023dreamgaussian} over DreamFusion have been introduced to address these issues. ProlificDreamer \cite{wang2023vsd} proposes variational score distillation (VSD) and produces high-fidelity texture results. Magic3D \cite{lin2023magic3d} adopts a coarse-to-fine strategy and utilizes DMTET \cite{shen2021deep} as the 3D representation to implement texture refinement through SDS loss. 
Despite yielding impressive progress, the appearance of their results is still unsatisfying, existing issues such as noise \cite{wang2023vsd}, lack of details \cite{lin2023magic3d, tang2023dreamgaussian}, multi-view inconsistency \cite{richardson2023texture, chen2023text2tex}. Moreover, these methods still face the problem of insufficient robustness and incorrect geometry. When it comes to avatar generation, these shortcomings can be more obvious and unacceptable.


\noindent {\bf Text-to-Avatar Generation.} To handle 3D avatar generation from text, extensive approaches \cite{hong2022avatarclip, dreamface, DreamAvatar, avatarcraft, dreamhuman, dreamwaltz} have been proposed. Avatar-CLIP \cite{hong2022avatarclip} sets the foundation by initializing human geometry with a shape VAE and employing CLIP to guide geometry and texture modeling. 
DreamAvatar \cite{DreamAvatar} and AvatarCraft \cite{avatarcraft} fulfill robust 3D avatar creation by integrating the human parametric model SMPL \cite{loper2023smpl} with pre-trained text-to-image diffusion models. DreamHuman \cite{dreamhuman} further introduces a camera zoom-in strategy to refine the local details of 6 important body regions. 
Recently, AvatarVerse \cite{Avatarverse} and a concurrent work \cite{pan2023avatarstudio} employ DensePose-conditioned ControlNet \cite{zhang2023adding} for SDS guidance to realize more stable avatar creation and pose control. Although these methods exhibit quite decent results, weak SDS guidance still hampers their performance in multi-view consistency and texture fidelity. 

\section{Methods}


In this section, we present DiffusionGAN3D, which boosts the performance of 3D domain adaptation and text-to-avatar by combining and taking advantage of 3D GANs and diffusion priors. Fig.~\ref{fig:framework} illustrates the overview of our framework. After introducing some preliminaries (Sec.~\ref{sec:preliminaries}), we first elaborate our designs in diffusion-guided 3D domain adaptation (Sec.~\ref{sec:domain_adaptation})
, where we propose a relative distance loss to resolve the problem of diversity loss caused by SDS. Then we extend this architecture and introduce a case-specific learnable triplane to fulfill 3D-GAN based text-to-avatar (Sec.~\ref{sec:text_to_avatar}).
Finally, we design a novel progressive texture refinement stage (Sec.~\ref{sec:texture_refinement}) 
to improve the detail and authenticity of the texture generated above.

\subsection{Preliminaries} \label{sec:preliminaries}

\noindent {\bf EG3D} \cite{chan2022efficient} is a SOTA 3D generative model, which employ triplane as 3D representation and integrate StyleGAN2 \cite{karras2020analyzing} generator with neural rendering \cite{mildenhall2021nerf} to achieve high quality 3D shapes and pose-controlled image synthesis. 
It is composed of (1) a mapping network that projects the input noise to the latent space $W$, (2) a triplane generator that synthesizes the triplane with the latent code as input, and (3) a decoder that includes a triplane decoder, volume rendering module and super-resolution module in sequence. Given a triplane and camera poses as input, the decoder generates high-resolution images with view consistency. 

\noindent {\bf Score Distillation Sampling (SDS)}, proposed by DreamFusion \cite{chan2022efficient}, utilizes a pre-trained diffusion model $\bm{\epsilon}_{\mathrm{\phi}}$ as prior for optimization of a 3D representation $\theta$. Given an image $\boldsymbol{x}=g(\theta)$ that is rendered from a differentiable model $g$, we add random noise $\bm{\epsilon}$ on $\boldsymbol{x}$ at noise level $t$ to obtain a noisy image $\boldsymbol{z_t}$. The SDS loss then optimizes $\theta$ by minimizing the difference between the predicted noise $\bm{\epsilon}_{\phi} (\boldsymbol{z_t}; \boldsymbol{y}, t)$ and the added noise $\bm{\epsilon}$, which can be presented as:
\begin{equation} \label{eq:sds}
\nabla_{\theta} L_{SDS}(\phi, g_{\theta}) = \mathbb{E}_{t, \epsilon} \left[ w_t \left( \boldsymbol{\epsilon}_{\phi}(\boldsymbol{z_t}; \boldsymbol{y}, t) - \boldsymbol{\epsilon} \right) \frac{\partial \boldsymbol{x}}{\partial \theta} \right], 
\end{equation}
where $\boldsymbol{y}$ indicates the text prompt and $w_t$ denotes a weighting function that depends on the noise level $t$.

\subsection{Diffusion-Guided 3D Domain Adaptation} \label{sec:domain_adaptation}

Due to the difficulties in obtaining high-quality pose-aware data and model training, adversarial learning methods for 3D domain adaptation mostly suffer from the issues of tedious data processing and mode collapse. To address that, we leverage diffusion models and adopt the SDS loss to implement transfer learning on an EG3D-based 3D GAN to achieve efficient 3D domain adaptation, as shown in Fig.~\ref{fig:framework}. 

Given a style code $\boldsymbol{w}$ generated from noise $\boldsymbol{z}\sim N(0, 1)$ through the fixed mapping network, we can obtain the triplane $\boldsymbol{T}$ and the image $\boldsymbol{x}$ rendered in a view controlled by the input camera parameters $\boldsymbol{c}$ using the triplane generator and decoder in sequence. Then SDS loss (Sec.~\ref{sec:preliminaries}) is applied on $\boldsymbol{x}$ to finetune the network. Different from DreamFusion which optimizes a NeRF network to implement single object generation, we shift the 3D generator with random noise and camera pose to achieve domain adaptation guided by text $\boldsymbol{y}$. 
During optimization, all parameters of the framework are frozen except the triplane generator. We find that the gradient provided by SDS loss is unstable and can be harmful to some other well-trained modules such as the super-resolution module. Besides, freezing the mapping network ensures that the latent code $\boldsymbol{w}$ lies in the same domain during training, which is a crucial feature that can be utilized in the diversity preserving of the 3D generator.

\begin{figure}[t]
  \centering
  \resizebox{0.88\linewidth}{!}{
   \includegraphics{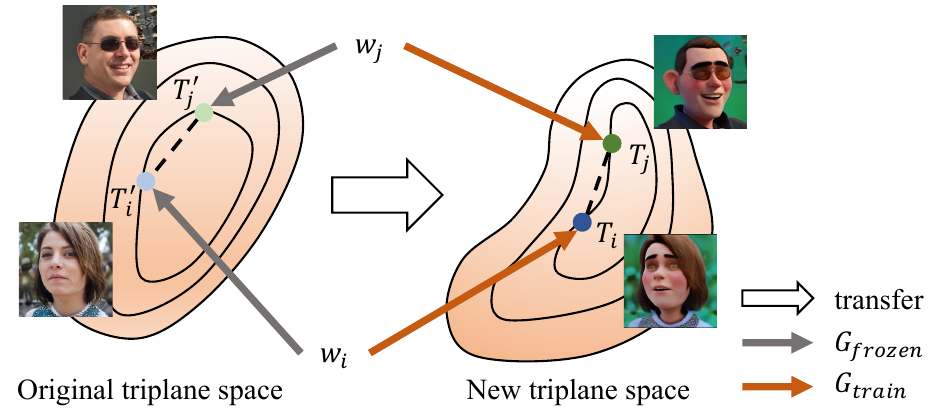} }
     \vspace{-5pt}
  \caption{An illustration of the relative distance loss.}
  \label{fig:relative_distance_loss}
  \vspace{-5pt}
\end{figure}

\noindent {\bf Relative Distance Loss.} The SDS loss provides diffusion priors and achieves text-guided domain adaptation of 3D GAN in an efficient way. However, it also brings the problem of diversity loss as illustrated in \cite{song2022diffusion}. To deal with that, \cite{song2022diffusion} proposes the directional regularizer to regularize the generator optimization process, which improves the diversity to a certain extent. However, it also limits the domain shifting, facing a trade-off between diversity and the degree of style transfer. To address this, we propose a relative distance loss. As shown in Fig.~\ref{fig:relative_distance_loss}, considering two style codes $\boldsymbol{w_i}$ and $\boldsymbol{w_j}$ which are mapping from two different noise $\boldsymbol{z_i}$ and $\boldsymbol{z_j}$, we project them into the original triplane domain ($\boldsymbol{T'_i}$, $\boldsymbol{T'_j}$) and the finetuned one ($\boldsymbol{T_i}$, $\boldsymbol{T_j}$) using a frozen triplane generator $G_{frozen}$ and the finetuned triplane generator $G_{train}$, respectively. Note that, since the mapping network is frozen during training in our framework, $\boldsymbol{T_i}$ and $\boldsymbol{T'_i}$ (same for $\boldsymbol{T_j}$ and $\boldsymbol{T'_j}$) share the same latent code and ought to be close in context. Thus, we model the relative distance of these two samples in triplane space and formulate the relative distance loss $L_{dis}$ as:

\begin{equation} \label{eq:relative_distance_loss}
L_{dis} = abs(\frac{\lvert \lvert \boldsymbol{T'_{i}} - \boldsymbol{T'_{j}} \rvert \rvert^2}{\lvert \lvert \boldsymbol{T_{i}} - \boldsymbol{T_{j}} \rvert \rvert^2} - 1).
\end{equation}

\noindent In this function, guided by the original network, the samples in the triplane space are forced to maintain distance from each other. This prevents the generator from collapsing to a fixed output pattern. 
Note that it only regularizes the relative distance between different samples while performing no limitation to the transfer of the triplane domain itself. Extensive experiments in Sec.~\ref{sec:experiments} demonstrate that the proposed relative distance loss effectively improves the generation diversity without impairing the degree of stylization.

\begin{figure}[t]
  \centering
  \resizebox{0.98\linewidth}{!}{
   \includegraphics{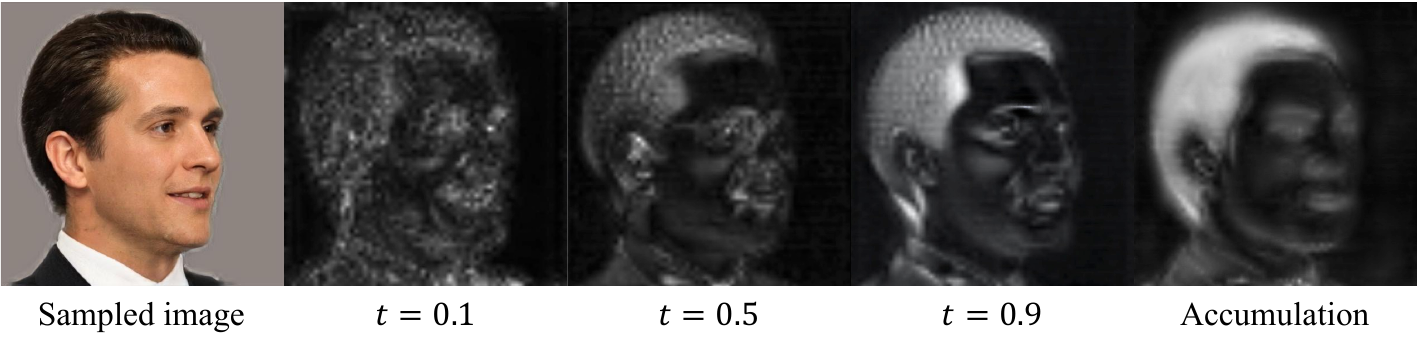} }
     \vspace{-5pt}
  \caption{Visualizations of the gradient response of SDS loss at different noise levels, given the text "a man with green hair".}
  \label{fig: diffusion_guided_rec_loss}
  \vspace{-12pt}
\end{figure}

\noindent {\bf Diffusion-guided Reconstruction Loss.} Despite the combination of SDS loss and the proposed relative distance loss is adequate for most domain adaptation tasks, it still fails to handle the local editing scenarios. A naive solution is to perform reconstruction loss between the rendered image and the one from the frozen network. However, it will also inhibit translation of the target region. Accordingly, we propose a diffusion-guided reconstruction loss especially for local editing, which aims to preserve non-target regions while performing 3D editing on the target region.  We found that the gradient of SDS loss has a certain correlation with the target area, especially when the noise level $t$ is large, as shown in Fig.~\ref{fig: diffusion_guided_rec_loss}. To this end, we design a diffusion-guided reconstruction loss $L_{diff}$ that can be presented as:

\begin{equation}
\gamma = abs(w_{t}(\boldsymbol{\epsilon}_{\phi}(\boldsymbol{z_t}; \boldsymbol{y}, t) - \boldsymbol{\epsilon})), 
\end{equation}
\begin{equation} \label{eq:diffusion_guided_loss}
L_{diff} = t \lvert \lvert (\boldsymbol{x} - \boldsymbol{x'}) \odot \left[ \boldsymbol{J} - h(\frac{\boldsymbol{\gamma}}{max(\boldsymbol{\gamma})}) \right] \rvert \rvert^2, 
\end{equation}

\noindent where $\boldsymbol{\gamma}$ is the absolute value of the gradient item in Eq.~\ref{eq:sds}, $h$ represents the averaging operation in the feature dimension, $\boldsymbol{J}$ is the matrix of ones having the same spatial dimensions as the output of $h$, $\boldsymbol{x'}$ denotes the output image of the frozen network under the same noise and camera parameters $\boldsymbol{x}$, $\odot$ indicates the Hadamard product. The latter item of the $\odot$ operation can be regarded as an adaptive mask indicating the non-target region. Compared with ordinary reconstruction loss, the proposed diffusion-guided reconstruction loss alleviates the transfer limitation of the target region. Although the gradient of SDS loss in a single iteration contains a lot of noise and is inadequate to serve as an accurate mask, it can also provide effective guidance for network learning with the accumulation of iterations as shown in Fig.~\ref{fig: diffusion_guided_rec_loss}. The ablation experiment in Sec.~\ref{sec:experiments} also proves its effectiveness.

To sum up, we can form the loss functions for normal domain adaptation and local editing scenario as $L_{adaptation} = L_{sds} + \lambda_1 L_{dis}$ and $L_{editing} = L_{sds} + \lambda_2 L_{diff}$, respectively, where $\lambda_1$ and $\lambda_2$ are the weighting coefficients.

\subsection{3D-GAN Based Text-to-Avatar} \label{sec:text_to_avatar}

Due to the lack of 3D priors 
, most text-to-3D methods cannot perform stable generation, suffering from issues such as Janua (multi-face) problem. To this end, we extend the framework proposed above and utilize the pre-trained 3D GAN as a strong base generator to achieve robust text-guided 3D avatar generation. As shown in Fig.~\ref{fig:framework}, we first implement latent searching to obtain the latent code that is contextually (gender, appearance, etc.) close to the text input.
Specifically, we sample $k$ noise $\boldsymbol{z_1},...,\boldsymbol{z_k}$ and select one single noise $\boldsymbol{z_i}$ that best fits the text description according to the CLIP loss between the corresponding images synthesized by the 3D GAN and the prompt. The CLIP loss is further used to finetune the mapping network individually to obtain the optimized latent code $\boldsymbol{w'}$ from $\boldsymbol{z_i}$. Then, $\boldsymbol{w'}$ is fixed during the following optimization process.

\noindent {\bf Case-specific learnable triplane.} One main challenge of the text-to-avatar task is how to model the highly variable geometry and texture. Introducing 3D GANs as the base generator provides strong priors and greatly improves stability. However, it also loses the flexibility of the simple NeRF network, showing limited generation capability. Accordingly, we introduce a case-specific learnable triplane $\boldsymbol{T_l}$ to enlarge the capacity of the network, as shown in Fig.~\ref{fig:framework}. Initialized with the value of 0, $\boldsymbol{T_l}$ is directly added to $\boldsymbol{T}$ as the input of subsequent modules. Thus, the trainable part of the network now includes the triplane generator $G_{train}$ and $\boldsymbol{T_l}$. The former achieves stable transformation, while the latter provides a more flexible 3D representation. Due to the high degree of freedom of $\boldsymbol{T_l}$ and the instability of SDS loss, optimizing $\boldsymbol{T_l}$ with SDS loss alone will bring a lot of noise, resulting in unsmooth results. To this end, we adopt the total variation loss \cite{johnson2016perceptual} and expand it to a multi-scale manner $L_{mstv}$ to regularize $\boldsymbol{T_l}$ and facilitate more smoothing results. In general, the loss function for text-to-avatar task can be presented as: $L_{avatar} = L_{sds} + \lambda_3 L_{mstv}$.

Note that, the proposed framework is only suitable for the generation of specific categories depending on the pre-trained 3D GAN, such as head (PanoHead \cite{an2023panohead}) and human body (AG3D \cite{dong2023ag3d}). Nevertheless, extensive experiments show that our framework can well adapt to avatar generation
with large domain gaps, benefiting from the strong 3D generator and the case-specific learnable triplane.

\subsection{Progressive Texture Refinement} \label{sec:texture_refinement}

\begin{figure}[t]
  \centering
  \resizebox{0.88\linewidth}{!}{
   \includegraphics{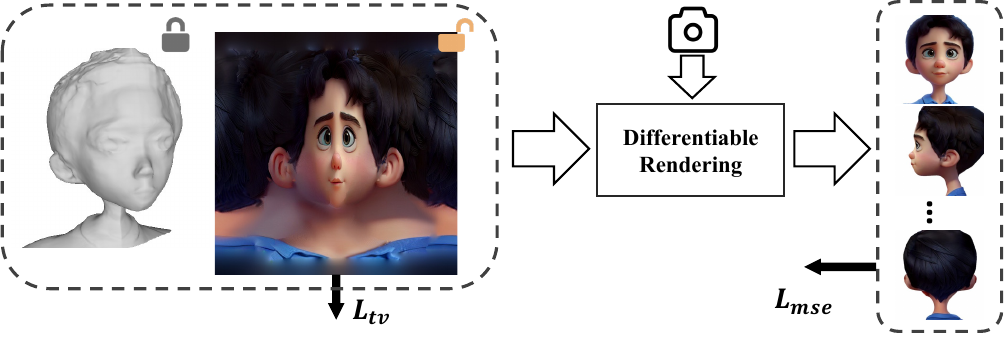} }
     \vspace{-5pt}
  \caption{The details of the proposed adaptive blend module.}
  \label{fig: adaptive_blend_module}
  \vspace{-10pt}
\end{figure} 

The SDS exhibits great promise in geometry modeling but also suffers from texture-related problems such as over-saturation and over-smoothing. 
How can we leverage the powerful 2D generation ability of diffusion models to improve the 3D textures? In this section, we propose a progressive texture refinement stage, which significantly enhances the texture quality of the results above through explicit texture modeling, as shown in Fig.~\ref{fig:framework}.

\noindent {\bf Adaptive Blend Module.} Given the implicit fields obtained from the first stage, we first implement volume rendering under uniformly selected $2k+2$ azimuths and $j$ elevations (we set the following $j$ to 1 for simplicity) to obtain multi-view images $\boldsymbol{x_i},..., \boldsymbol{x_{2k+1}}$. Then the canny maps and depth maps of these images are extracted for the following image translation. Meanwhile, we perform marching cube \cite{lorensen1998marching} and the UV unwrapping \cite{xatlas} algorithm to obtain the explicit mesh $\boldsymbol{M}$ and corresponding UV coordinates (in head generation, we utilize cylinder unwrapping for better visualization). Furthermore, we design an adaptive blend module to project the multi-view renderings back into a texture map through differentiable rendering. Specifically, as shown in Fig.~\ref{fig: adaptive_blend_module}, the multi-view reconstruction loss $L_{mse}$ and total variation loss $L_{tv}$ are adopted to optimize the texture map that is initialized with zeros. Compared to directly implementing back-projection, the proposed adaptive blending module produces smoother and more natural textures in spliced areas of different images without compromising texture quality. This optimized UV texture $\boldsymbol{U_0}$ serves as an initialization for the following texture refinement stage.

\noindent {\bf Progressive Refinement.}
Since we have already obtained the explicit mesh and the multi-view renderings, a natural idea is to perform image-to-image on the multi-view renderings using diffusion models to optimize the texture. However, it neglects that the diffusion model cannot guarantee the consistency of image translation between different views, which may result in discontinuous texture. To this end, we introduce a progressive inpainting strategy to address this issue. Firstly, we employ a pre-trained text-to-image diffusion model and ControlNets \cite{zhang2023adding} to implement image-to-image translation guided by the prompt $\boldsymbol{y}$ on the front-view image $\boldsymbol{r_0}$ that is rendered from $\boldsymbol{M}$ and $\boldsymbol{U_0}$. The canny and depth extracted above are introduced to ensure the alignment between $\boldsymbol{r_0}$ and the resulting image $\boldsymbol{r'_0}$. Then we can obtain the partially refined texture map $\boldsymbol{U_1}$ by projecting $\boldsymbol{r'_0}$ into $\boldsymbol{U_0}$. Next, we rotate the mesh coupled with $\boldsymbol{T_1}$ (or change the camera view) and render a new image $\boldsymbol{r_1}$, which is refined again with the diffusion model to get $\boldsymbol{r'_1}$ and $\boldsymbol{U_2}$. Differently, instead of image-to-image, we apply inpainting on $\boldsymbol{r_1}$ with mask $\boldsymbol{m_1}$ in this translation, which maintains the refined region and thus improves the texture consistency between the adjacent views. Note that the masks $\boldsymbol{m_1}, ..., \boldsymbol{m_{2k+1}}$ indicate the unrefined regions and are dilated to facilitate smoother results in inpainting. Through progressively performing rotation and inpainting, we manage to obtain consistent multi-view images $\boldsymbol{r'_0}, ..., \boldsymbol{r'_{2k+1}}$ that are refined by the diffusion model. Finally, we apply the adaptive blend module again on the refined images to yield the final texture. By implementing refinement on the explicit texture, the proposed stage significantly improves the texture quality in an efficient way.

\section{Experiments} \label{sec:experiments}

\subsection{Implementation Details}

Our framework is built on an EG3D-based model in the first stage. Specifically, we implement 3D domain adaptation on PanoHead, EG3D-FFHQ, and EG3D-AFHQ for head, face, and cat, respectively. For text-to-avatar tasks, PanoHead and AG3D are adopted as the base generators for head and body generation. We employ StableDiffusion v2.1 as our pre-trained text-to-image model. 
In the texture refinement stage, StableDiffusion v1.5 coupled with ControlNets are utilized to implement image-to-image and inpainting.
More details about the parameters and training setting are specified in supplementary materials.



\subsection{Qualitative Comparison}

\begin{figure*}[t]
   \setlength{\belowcaptionskip}{-0.2cm}
   \setlength{\abovecaptionskip}{0.1cm}
\centerline{\includegraphics[scale=0.53]{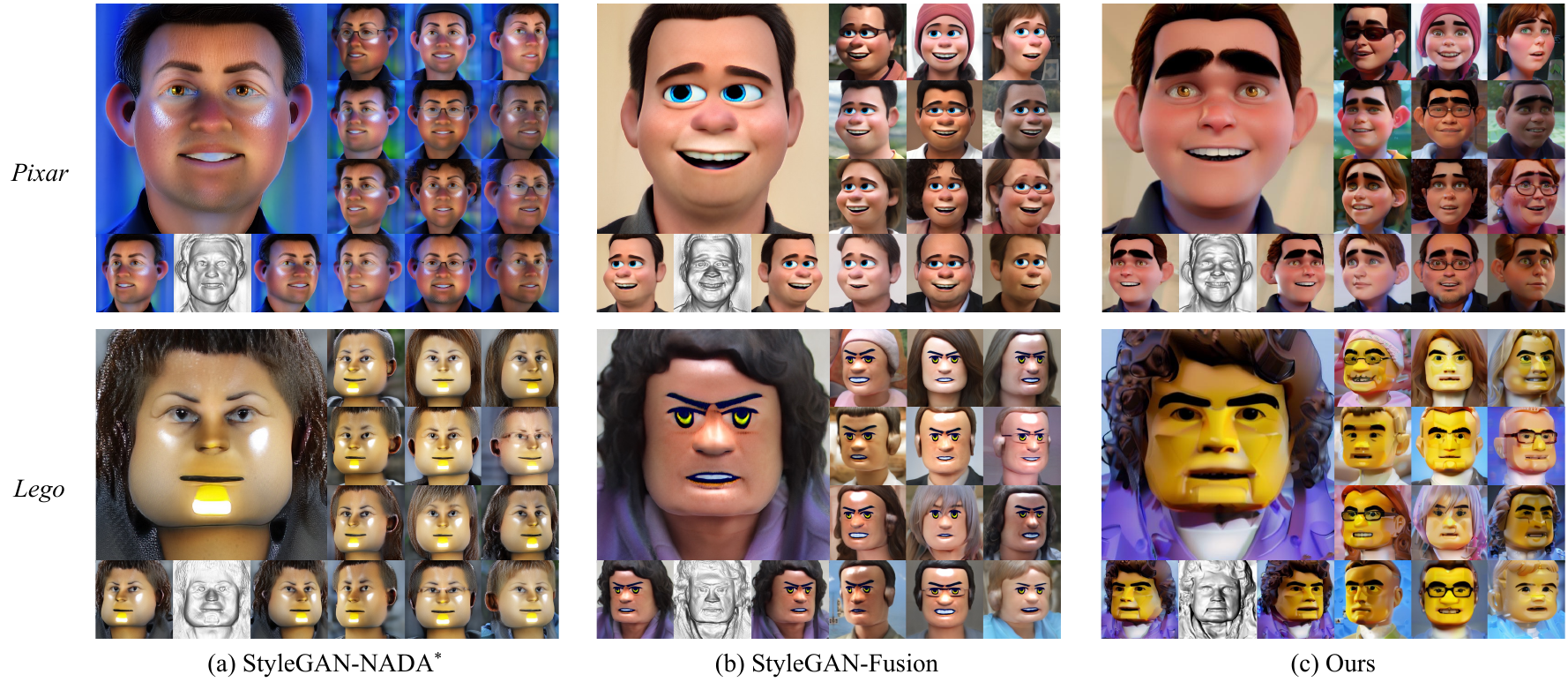}}
\caption{The qualitative comparisons on 3D domain adaptation (applied on EG3D-FFHQ \cite{chan2022efficient}).}
\label{fig:comparison_domain_adaptation}
\end{figure*}

For 3D Domain adaptation, we evaluate our model with two powerful baselines: StyleGAN-NADA* \cite{gal2021stylegan} and StyleGAN-Fusion \cite{song2022diffusion} for text-guided domain adaptation of 3D GANs, where * indicates the extension of the method to 3D models. For a fair comparison, we use the same prompts as guidance for all the methods. Besides, the visualization results of different methods are sampled from the same random noise.
As shown in Fig.~\ref{fig:comparison_domain_adaptation}, the naive extension of StyleGAN-NADA* for EG3D exhibits poor results in terms of diversity and image quality.
StyleGAN-Fusion achieves decent 3D domain adaptation and exhibits a certain diversity. However, the proposed regularizer of StyleGAN-Fusion also hinders itself from large-gap domain transfer, resulting in a trade-off between the degree of stylization and diversity. As Fig.~\ref{fig:comparison_domain_adaptation} shows that the generated faces of StyleGAN-Fusion lack diversity and details, and the hair and clothes suffer from inadequate stylization. In contrast, our method exhibits superior performance in diversity, image quality, and text-image correspondence.

\begin{figure*}[t]
   \setlength{\belowcaptionskip}{-0.2cm}
   \setlength{\abovecaptionskip}{0.1cm}
\centerline{\includegraphics[scale=0.53]{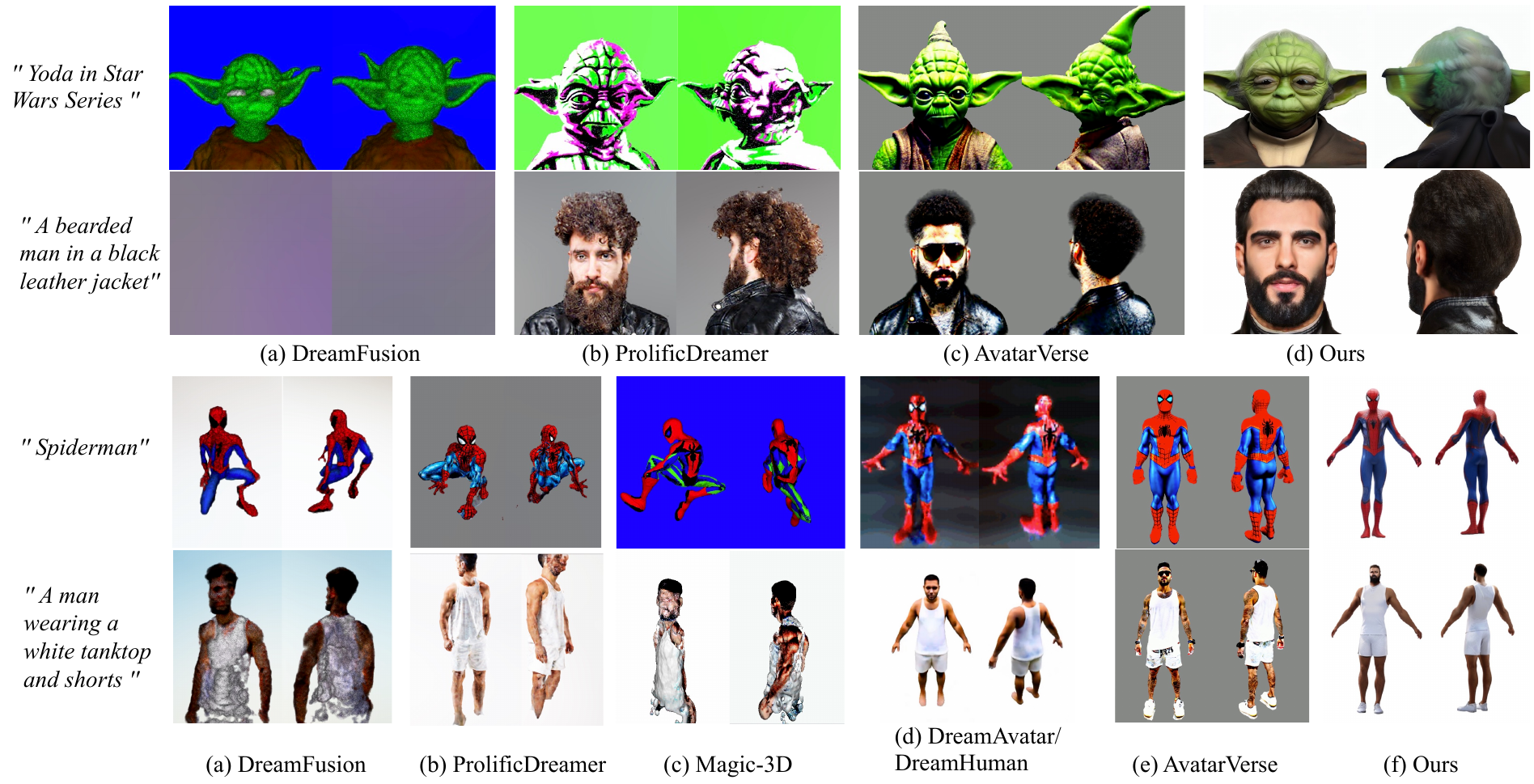}}
\caption{Visual comparisons on text-to-avatar task. The first two rows are the results of ‘head' and the rest are the results of ‘body'.}
\label{fig:comparison_text}
\end{figure*}

For text-to-avatar task, We present qualitative comparisons with several general text-to-3D methods (DreamFusion \cite{poole2022dreamfusion}, ProlificDreamer \cite{wang2023vsd}, Magic-3D \cite{lin2023magic3d}) and avatar generation methods (DreamAvatar \cite{DreamAvatar}, DreamHuman \cite{dreamhuman}, AvatarVerse \cite{Avatarverse}). The former three methods are implemented using the official code and the results of the rest methods are obtained directly from their project pages. As shown in Fig.~\ref{fig:comparison_text}, DreamFusion shows inferior performance in avatar generation, suffering from over-saturation, Janus (multi-face) problem, and incorrect body parts. ProlificDreamer and Magic-3D improve the texture fidelity to some extent but still face the problem of inaccurate and unsmooth geometry. Taking advantage of the human priors obtained from the SMPL model or DensePose, these text-to-avatar methods achieve stable and high-quality avatar generation. However, due to that the SDS loss requires a high CFG \cite{ho2022classifier} value during optimization, the texture fidelity and authenticity of their results are still unsatisfying.
In comparison, the proposed method achieves stable and high-fidelity avatar generation simultaneously, making full use of the 3D GANs and diffusion priors. Please refer to the supplementary materials for more comparisons.

\subsection{Quantitative Comparison}

We quantitatively evaluate the above baselines and our method on 3D domain adaptation through FID \cite{heusel2017gans} comparison and user study. Specifically, all methods are employed to conduct domain adaptation on EG3D-face and EG3D-cat with both four text prompts, respectively. For each text prompt, we utilize the text-to-image diffusion model to generate 2000 images with different random seeds as the ground truth for FID calculation. In the user study, 12 volunteers were invited to rate each finetuned model from 1 to 5 based on three dimensions: text-image correspondence, image quality, and diversity. As shown in Table ~\ref{tab:comp_domain_adaptation}, the proposed methods achieve lower FID scores than other baselines, which indicates superior image fidelity. Meanwhile, the user study demonstrates that our method outperforms the other two methods, especially in terms of image quality and diversity.

For text-to-avatar, we also conducted a user study for quantitative comparison. Since AvatarVerse and DreamAvatar have not released their code yet, while DreamHuman provided extensive results on the project page. So we compare our method with DreamHuman for full-body generation. Besides, DreamFusion, ProlificDreamer, and Magic3D are involved in the comparison of head (10 prompts) and full-body (10 prompts) generation both. We request the 12 volunteers to vote for their favorite results based on texture and geometry quality, where all the results are presented as rendered rotating videos. The final rates presented in Table ~\ref{tab:comp_text_to_avatar} show that the proposed method performs favorably against the other approaches.

\begin{table}[ht]
 \centering
  \caption{Quantitative comparison on 3D domain adaptation task. } %
    \vspace{-5pt}
\footnotesize
\resizebox{1.00\linewidth}{!}{
\begin{tabular}{l|c|ccc}
    \toprule
\multirow{2}{*}{Methods} & Metric & \multicolumn{3}{c}{User Study}   \\ \cline{2-5}

   & FID $\downarrow$ & text-corr $\uparrow$ & quality $\uparrow$  & diversity $\uparrow$  \\
    \midrule
StyleGAN-NADA*     &  136.2 & 2.619 & 2.257 & 1.756     \\
StyleGAN-Fusion     &  53.6 & 3.465 & 3.255 & 2.978     \\
Ours     &  \textbf{28.5} & \textbf{3.725} & \textbf{3.758} & \textbf{3.416}   \\
    \midrule

\end{tabular}}
  \label{tab:comp_domain_adaptation}
  \vspace{-5pt}
\end{table}

\begin{table}[ht]
 \centering
  \caption{User preference on text-to-avatar generation. } %
    \vspace{-5pt}
\footnotesize
\resizebox{1.00\linewidth}{!}{
\begin{tabular}{c|c|c|c|c|c}
    \toprule
   & DreamFusion & ProlificDreamer & Magic3D  & DreamHuman & Ours  \\
    \midrule
head     &  1.1\% & 11.7\% & 6.7\% & N.A. &  \textbf{80.5\%}   \\
body     &  0.6\% & 8.3\% & 5.6\% & 18.9\% &  \textbf{66.6\%}     \\
    \midrule

\end{tabular}}
  \label{tab:comp_text_to_avatar}
  \vspace{-10pt}
\end{table}

\subsection{Ablation Study}

\begin{figure}[t]
  \centering
  \resizebox{0.98\linewidth}{!}{
   \includegraphics{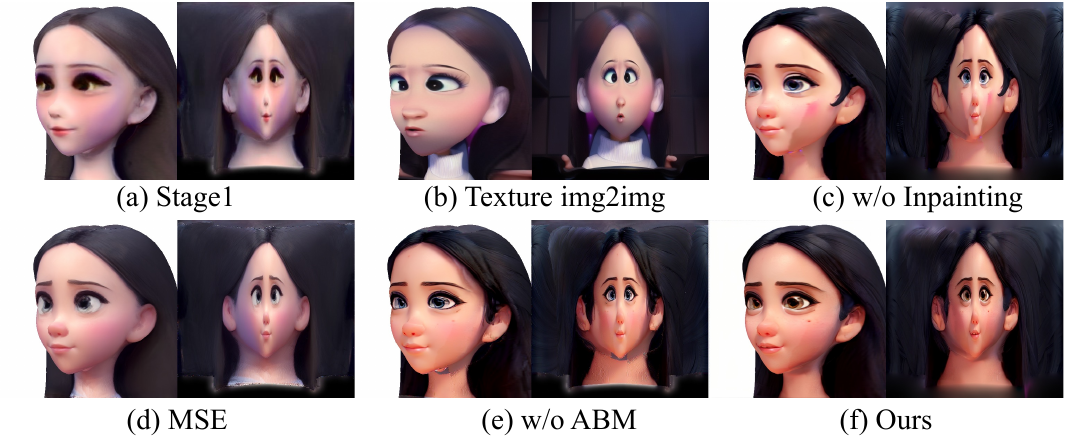} }
     \vspace{-3pt}
  \caption{Ablation study of the texture refinement.}
  \label{fig: ablation_texture_refinement}
  \vspace{-2pt}
\end{figure}

\noindent {\bf On progressive texture refinement.} Since we utilize cylinder unwrapping for head texture refinement, a naive idea is to conduct image-to-image on the UV texture directly to refine it. However, the result in Fig.~\ref{fig: ablation_texture_refinement} (b) shows that this method tends to yield misaligned texture, let alone be applied to fragmented texture maps. We also attempt to replace all the inpainting operations with image-to-image translation, and the results in Fig.~\ref{fig: ablation_texture_refinement} (c) show that this will cause the discontinuity problem. The refining strategy proposed in \cite{tang2023dreamgaussian} is also compared, where texture is progressively optimized using MSE loss between the randomly rendered images and the corresponding image-to-image results. The results in Fig.~\ref{fig: ablation_texture_refinement} (d) show that it fails to generate high-frequency details. The comparison between (e) and (f) in Fig.~\ref{fig: ablation_texture_refinement} proves the effectiveness of the proposed adaptive blend module (ABM) in smoothing the texture splicing region. By contrast, the proposed progressive texture refinement strategy significantly improves the texture quality.

\begin{figure}[t]
  \centering
  \resizebox{0.98\linewidth}{!}{
   \includegraphics{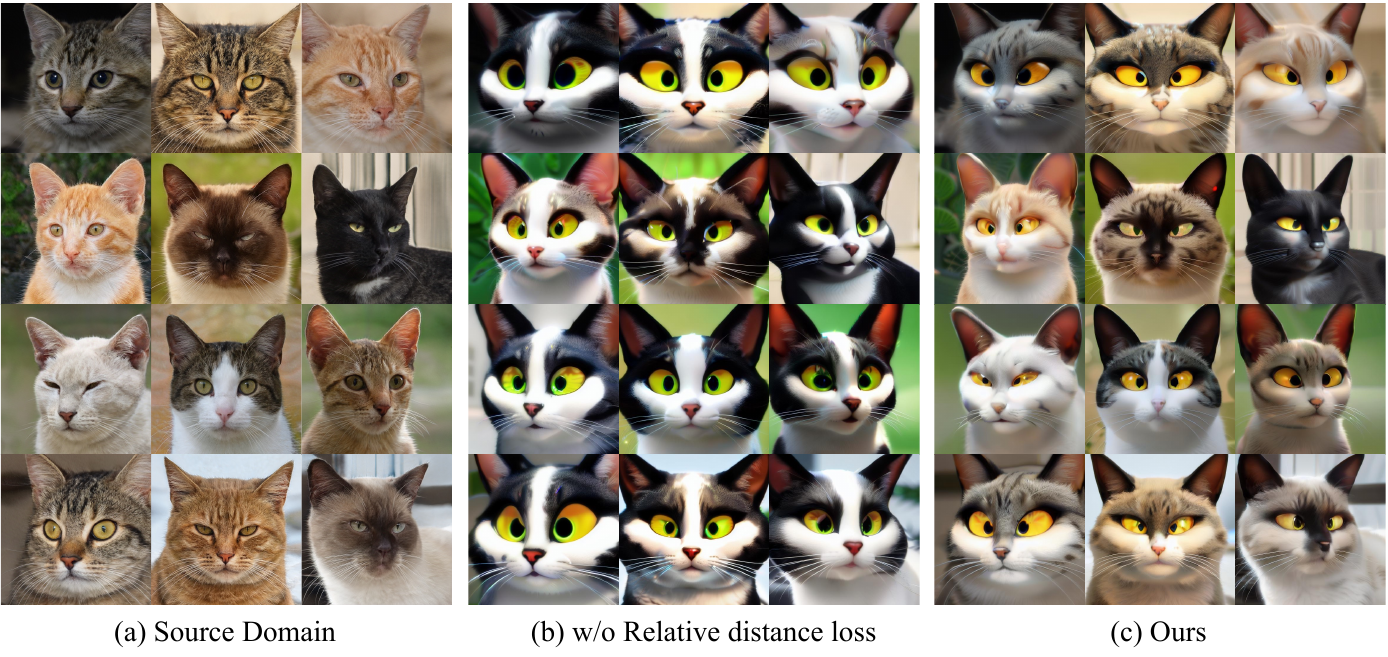} }
     \vspace{-5pt}
  \caption{Ablation study of the relative distance loss.}
  \label{fig: ablation_diversity}
  \vspace{-2pt}
\end{figure}

\noindent {\bf On relative distance loss.} As shown in Fig.~\ref{fig: ablation_diversity}, if adopting SDS loss alone for domain adaptation, the generator will tend to collapse to a fixed output pattern, losing its original diversity. In contrast, the proposed relative distance loss effectively preserves the diversity of the generator without sacrificing the stylization degree.

\begin{figure}[t]
  \centering
  \resizebox{0.98\linewidth}{!}{
   \includegraphics{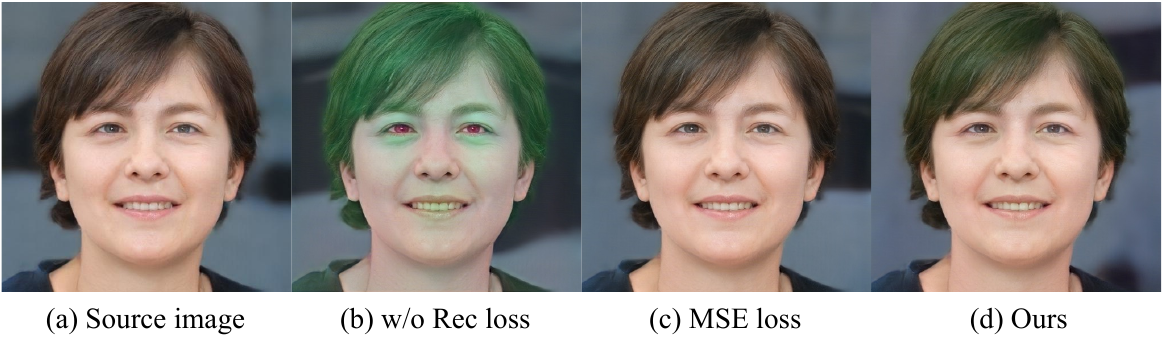} }
  \caption{Ablation study of the diffusion guided reconstruction loss. The ToRGB module in EG3D is trained together with $G_{train}$. The input text is “a close-up of a woman with green hair". }
  \label{fig: ablation_editing}
\end{figure}

\noindent {\bf On diffusion-guided reconstruction loss.} The results in Fig~\ref{fig: ablation_editing} show that the SDS loss tends to perform global transfer. Regular reconstruction loss helps maintain the whole structure, but also stem the translation of the target area. By contrast, the model trained with our diffusion-guided reconstruction loss achieves proper editing.

\begin{figure}[t]
  \centering
  \resizebox{0.98\linewidth}{!}{
   \includegraphics{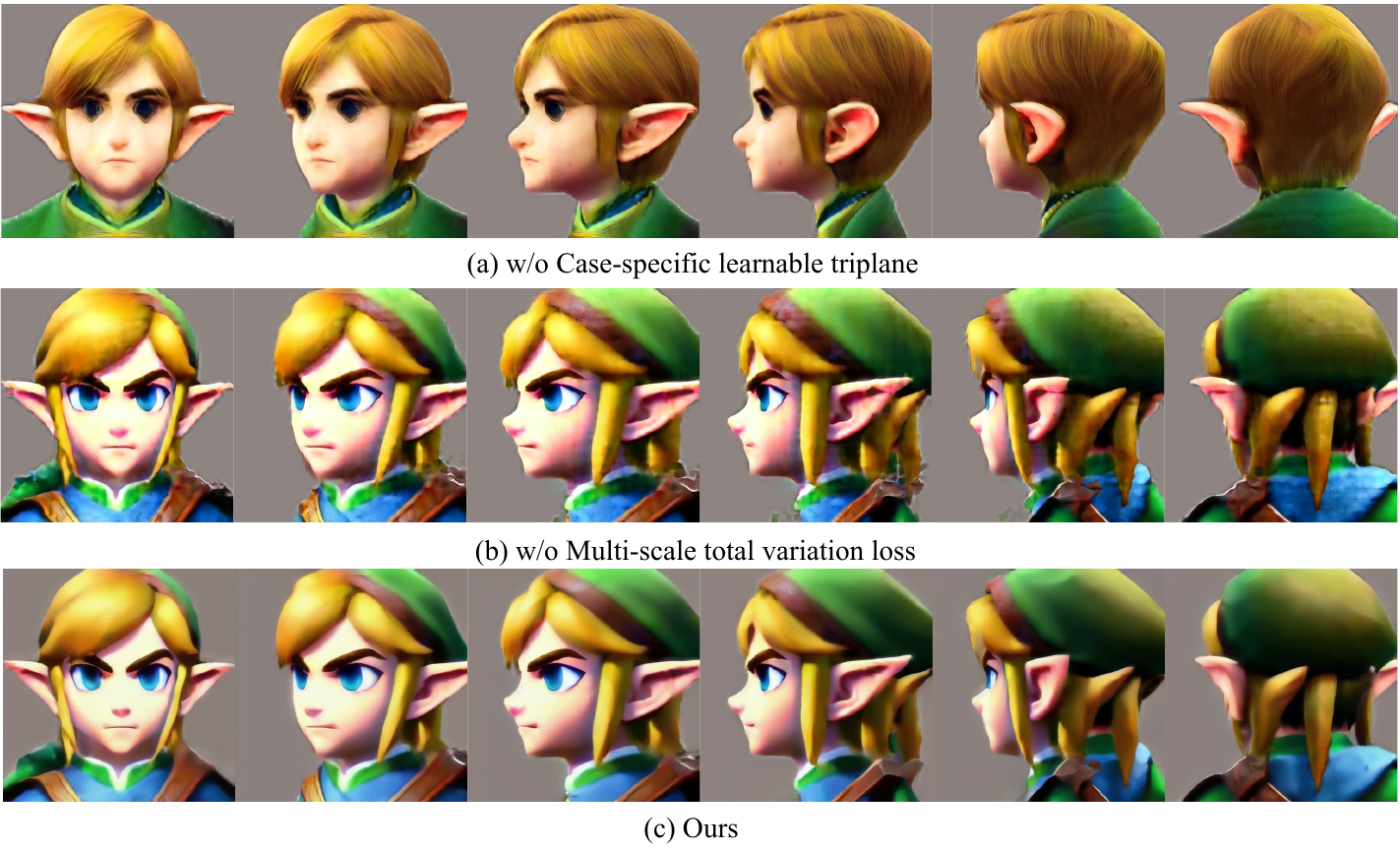} }
     \vspace{-5pt}
  \caption{Ablation study toward the case-specific learnable triplane and the multi-scale total variation loss.}
  \label{fig: ablation_learnable_triplane}
  \vspace{-2pt}
\end{figure}

\noindent {\bf On additional learnable triplane.} To prove the necessity of the proposed case-specific learnable triplane, we finetune the network with SDS loss without adding it, given a challenging prompt: "Link in Zelda". The results in the first row of Fig.~\ref{fig: ablation_learnable_triplane} reveal that the network is optimized in the right direction but fails to reach the precise point. By contrast, the network adding the learnable triplane exhibits accurate generation (second row in Fig.~\ref{fig: ablation_learnable_triplane}). Furthermore, the introduced multi-scale total variation loss $L_{mstv}$ on the triplane facilitates more smooth results.

\subsection{Applications and Limitations}
Due to the page limitation, we will introduce the application of DiffusionGAN3D on real images and specify the limitations of our methods in the supplementary materials.

\section{Conclusion}

In this paper, we propose a novel two-stage framework DiffusionGAN3D, which boosts text-guided 3D domain adaptation and avatar generation by combining the 3D GANs and diffusion priors. Specifically, we integrate the pre-trained 3D generative models (e.g., EG3D) with the text-to-image diffusion models. The former, in our framework, set a strong foundation for text-to-avatar, enabling stable and high-quality 3D avatar generation. In return, the latter provides informative direction for 3D GANs to evolve, which facilitates the text-guided domain adaptation of 3D GANs in an efficient way. Moreover, we introduce a progressive texture refinement stage, which significantly enhances the texture quality of the generation results. Extensive experiments demonstrate that the proposed framework achieves excellent results in both domain adaptation and text-to-avatar tasks, outperforming existing methods in terms of generation quality and efficiency.

\clearpage

{
    \small
    \bibliographystyle{ieeenat_fullname}
    \bibliography{main}

\begin{thebibliography}{56}
\providecommand{\natexlab}[1]{#1}
\providecommand{\url}[1]{\texttt{#1}}
\expandafter\ifx\csname urlstyle\endcsname\relax
  \providecommand{\doi}[1]{doi: #1}\else
  \providecommand{\doi}{doi: \begingroup \urlstyle{rm}\Url}\fi

\bibitem[xat()]{xatlas}
Jonathan young. xatlas, 2021.
\newblock \url{https://triplegangers.com/}.

\bibitem[Abdal et~al.(2023)Abdal, Lee, Zhu, Chai, Siarohin, Wonka, and Tulyakov]{abdal20233davatargan}
Rameen Abdal, Hsin-Ying Lee, Peihao Zhu, Menglei Chai, Aliaksandr Siarohin, Peter Wonka, and Sergey Tulyakov.
\newblock 3davatargan: Bridging domains for personalized editable avatars.
\newblock In \emph{Proceedings of the IEEE/CVF Conference on Computer Vision and Pattern Recognition}, pages 4552--4562, 2023.

\bibitem[Alanov et~al.(2022)Alanov, Titov, and Vetrov]{alanov2022hyperdomainnet}
Aibek Alanov, Vadim Titov, and Dmitry~P Vetrov.
\newblock Hyperdomainnet: Universal domain adaptation for generative adversarial networks.
\newblock \emph{Advances in Neural Information Processing Systems}, 35:\penalty0 29414--29426, 2022.

\bibitem[An et~al.(2023)An, Xu, Shi, Song, Ogras, and Luo]{an2023panohead}
Sizhe An, Hongyi Xu, Yichun Shi, Guoxian Song, Umit~Y Ogras, and Linjie Luo.
\newblock Panohead: Geometry-aware 3d full-head synthesis in 360deg.
\newblock In \emph{Proceedings of the IEEE/CVF Conference on Computer Vision and Pattern Recognition}, pages 20950--20959, 2023.

\bibitem[Cao et~al.(2023)Cao, Cao, Han, Shan, and Wong]{DreamAvatar}
Yukang Cao, YanPei Cao, Kai Han, Ying Shan, and Kwan-Yee~K. Wong.
\newblock Dreamavatar: Text-and-shape guided 3d human avatar generation via diffusion models.
\newblock \emph{arXiv preprint arXiv:2304.00916}, 2023.

\bibitem[Chan et~al.(2021)Chan, Monteiro, Kellnhofer, Wu, and Wetzstein]{chan2021pi}
Eric~R Chan, Marco Monteiro, Petr Kellnhofer, Jiajun Wu, and Gordon Wetzstein.
\newblock pi-gan: Periodic implicit generative adversarial networks for 3d-aware image synthesis.
\newblock In \emph{Proceedings of the IEEE/CVF conference on computer vision and pattern recognition}, pages 5799--5809, 2021.

\bibitem[Chan et~al.(2022)Chan, Lin, Chan, Nagano, Pan, De~Mello, Gallo, Guibas, Tremblay, Khamis, et~al.]{chan2022efficient}
Eric~R Chan, Connor~Z Lin, Matthew~A Chan, Koki Nagano, Boxiao Pan, Shalini De~Mello, Orazio Gallo, Leonidas~J Guibas, Jonathan Tremblay, Sameh Khamis, et~al.
\newblock Efficient geometry-aware 3d generative adversarial networks.
\newblock In \emph{Proceedings of the IEEE/CVF Conference on Computer Vision and Pattern Recognition}, pages 16123--16133, 2022.

\bibitem[Chen et~al.(2023)Chen, Siddiqui, Lee, Tulyakov, and Nie{\ss}ner]{chen2023text2tex}
Dave~Zhenyu Chen, Yawar Siddiqui, Hsin-Ying Lee, Sergey Tulyakov, and Matthias Nie{\ss}ner.
\newblock Text2tex: Text-driven texture synthesis via diffusion models.
\newblock \emph{arXiv preprint arXiv:2303.11396}, 2023.

\bibitem[Dong et~al.(2023)Dong, Chen, Yang, Black, Hilliges, and Geiger]{dong2023ag3d}
Zijian Dong, Xu Chen, Jinlong Yang, Michael~J Black, Otmar Hilliges, and Andreas Geiger.
\newblock Ag3d: Learning to generate 3d avatars from 2d image collections.
\newblock \emph{arXiv preprint arXiv:2305.02312}, 2023.

\bibitem[et~al(2022)]{ramesh2022}
Aditya~Ramesh et al.
\newblock Hierarchical text-conditional image generation with clip latents, 2022.

\bibitem[Gadelha et~al.(2017)Gadelha, Maji, and Wang]{gadelha20173d}
Matheus Gadelha, Subhransu Maji, and Rui Wang.
\newblock 3d shape induction from 2d views of multiple objects.
\newblock In \emph{2017 International Conference on 3D Vision (3DV)}, pages 402--411. IEEE, 2017.

\bibitem[Gal et~al.(2021)Gal, Patashnik, Maron, Chechik, and Cohen-Or]{gal2021stylegan}
Rinon Gal, Or Patashnik, Haggai Maron, Gal Chechik, and Daniel Cohen-Or.
\newblock Stylegan-nada: Clip-guided domain adaptation of image generators.
\newblock \emph{arXiv preprint arXiv:2108.00946}, 2021.

\bibitem[Gu et~al.(2021)Gu, Liu, Wang, and Theobalt]{gu2021stylenerf}
Jiatao Gu, Lingjie Liu, Peng Wang, and Christian Theobalt.
\newblock Stylenerf: A style-based 3d-aware generator for high-resolution image synthesis.
\newblock \emph{arXiv preprint arXiv:2110.08985}, 2021.

\bibitem[Henzler et~al.(2019)Henzler, Mitra, and Ritschel]{henzler2019escaping}
Philipp Henzler, Niloy~J Mitra, and Tobias Ritschel.
\newblock Escaping plato's cave: 3d shape from adversarial rendering.
\newblock In \emph{Proceedings of the IEEE/CVF International Conference on Computer Vision}, pages 9984--9993, 2019.

\bibitem[Heusel et~al.(2017)Heusel, Ramsauer, Unterthiner, Nessler, and Hochreiter]{heusel2017gans}
Martin Heusel, Hubert Ramsauer, Thomas Unterthiner, Bernhard Nessler, and Sepp Hochreiter.
\newblock Gans trained by a two time-scale update rule converge to a local nash equilibrium.
\newblock \emph{Advances in neural information processing systems}, 30, 2017.

\bibitem[Ho and Salimans(2022)]{ho2022classifier}
Jonathan Ho and Tim Salimans.
\newblock Classifier-free diffusion guidance.
\newblock \emph{arXiv preprint arXiv:2207.12598}, 2022.

\bibitem[Hong et~al.(2022{\natexlab{a}})Hong, Chen, Lan, Pan, and Liu]{hong2022eva3d}
Fangzhou Hong, Zhaoxi Chen, Yushi Lan, Liang Pan, and Ziwei Liu.
\newblock Eva3d: Compositional 3d human generation from 2d image collections.
\newblock \emph{arXiv preprint arXiv:2210.04888}, 2022{\natexlab{a}}.

\bibitem[Hong et~al.(2022{\natexlab{b}})Hong, Zhang, Pan, Cai, Yang, and Liu]{hong2022avatarclip}
Fangzhou Hong, Mingyuan Zhang, Liang Pan, Zhongang Cai, Lei Yang, and Ziwei Liu.
\newblock Avatarclip: Zero-shot text-driven generation and animation of 3d avatars.
\newblock \emph{arXiv preprint arXiv:2205.08535}, 2022{\natexlab{b}}.

\bibitem[Huang et~al.(2023)Huang, Wang, Zeng, Cao, Qi, Shi, Zha, and Zhang]{dreamwaltz}
Yukun Huang, Jianan Wang, Ailing Zeng, He Cao, Xianbiao Qi, Yukai Shi, Zheng-Jun Zha, and Lei Zhang.
\newblock Dreamwaltz: Make a scene with complex 3d animatable avatars.
\newblock \emph{arXiv preprint arXiv:2305.12529}, 2023.

\bibitem[Jain et~al.(2022)Jain, Mildenhall, Barron, Abbeel, and Poole]{2022dreamfields}
Ajay Jain, Ben Mildenhall, Jonathan~T Barron, Pieter Abbeel, and Ben Poole.
\newblock Zero-shot text-guided object generation with dream fields.
\newblock In \emph{Proceedings of the IEEE/CVF Conference on Computer Vision and Pattern Recognition}, pages 867--876, 2022.

\bibitem[Jiang et~al.(2023)Jiang, Wang, Zhang, Chai, He, Chen, and Liao]{avatarcraft}
Ruixiang Jiang, Can Wang, Jingbo Zhang, Menglei Chai, Mingming He, Dongdong Chen, and Jing Liao.
\newblock Avatarcraft: Transforming text into neural human avatars with parameterized shape and pose control.
\newblock \emph{arXiv preprint arXiv:2303.17606}, 2023.

\bibitem[Jin et~al.(2022)Jin, Ryu, Kim, Baek, and Cho]{jin2022dr}
Wonjoon Jin, Nuri Ryu, Geonung Kim, Seung-Hwan Baek, and Sunghyun Cho.
\newblock Dr.3d: Adapting 3d gans to artistic drawings.
\newblock In \emph{SIGGRAPH Asia 2022 Conference Papers}, pages 1--8, 2022.

\bibitem[Johnson et~al.(2016)Johnson, Alahi, and Fei-Fei]{johnson2016perceptual}
Justin Johnson, Alexandre Alahi, and Li Fei-Fei.
\newblock Perceptual losses for real-time style transfer and super-resolution.
\newblock In \emph{European conference on computer vision}, pages 694--711. Springer, 2016.

\bibitem[Karras et~al.(2020)Karras, Laine, Aittala, Hellsten, Lehtinen, and Aila]{karras2020analyzing}
Tero Karras, Samuli Laine, Miika Aittala, Janne Hellsten, Jaakko Lehtinen, and Timo Aila.
\newblock Analyzing and improving the image quality of stylegan.
\newblock In \emph{Proceedings of the IEEE/CVF conference on computer vision and pattern recognition}, pages 8110--8119, 2020.

\bibitem[Kim and Chun(2023)]{kim2023datid}
Gwanghyun Kim and Se~Young Chun.
\newblock Datid-3d: Diversity-preserved domain adaptation using text-to-image diffusion for 3d generative model.
\newblock In \emph{Proceedings of the IEEE/CVF Conference on Computer Vision and Pattern Recognition}, pages 14203--14213, 2023.

\bibitem[Kim et~al.(2023)Kim, Jang, and Chun]{kim2023podia}
Gwanghyun Kim, Ji~Ha Jang, and Se~Young Chun.
\newblock Podia-3d: Domain adaptation of 3d generative model across large domain gap using pose-preserved text-to-image diffusion.
\newblock In \emph{Proceedings of the IEEE/CVF International Conference on Computer Vision}, pages 22603--22612, 2023.

\bibitem[Kolotouros et~al.(2023)Kolotouros, Alldieck, Zanfir, Bazavan, Fieraru, and Sminchisescu]{dreamhuman}
Nikos Kolotouros, Thiemo Alldieck, Andrei Zanfir, Eduard~Gabriel Bazavan, Mihai Fieraru, and Cristian Sminchisescu.
\newblock Dreamhuman: Animatable 3d avatars from text.
\newblock \emph{arXiv preprint arXiv:2306.09329}, 2023.

\bibitem[Lan et~al.(2023)Lan, Meng, Yang, Loy, and Dai]{lan2023self}
Yushi Lan, Xuyi Meng, Shuai Yang, Chen~Change Loy, and Bo Dai.
\newblock Self-supervised geometry-aware encoder for style-based 3d gan inversion.
\newblock In \emph{Proceedings of the IEEE/CVF Conference on Computer Vision and Pattern Recognition}, pages 20940--20949, 2023.

\bibitem[Liao et~al.(2020)Liao, Schwarz, Mescheder, and Geiger]{liao2020towards}
Yiyi Liao, Katja Schwarz, Lars Mescheder, and Andreas Geiger.
\newblock Towards unsupervised learning of generative models for 3d controllable image synthesis.
\newblock In \emph{Proceedings of the IEEE/CVF conference on computer vision and pattern recognition}, pages 5871--5880, 2020.

\bibitem[Lin et~al.(2023)Lin, Gao, Tang, Takikawa, Zeng, Huang, Kreis, Fidler, Liu, and Lin]{lin2023magic3d}
Chen-Hsuan Lin, Jun Gao, Luming Tang, Towaki Takikawa, Xiaohui Zeng, Xun Huang, Karsten Kreis, Sanja Fidler, Ming-Yu Liu, and Tsung-Yi Lin.
\newblock Magic3d: High-resolution text-to-3d content creation.
\newblock In \emph{Proceedings of the IEEE/CVF Conference on Computer Vision and Pattern Recognition}, pages 300--309, 2023.

\bibitem[Loper et~al.(2023)Loper, Mahmood, Romero, Pons-Moll, and Black]{loper2023smpl}
Matthew Loper, Naureen Mahmood, Javier Romero, Gerard Pons-Moll, and Michael~J Black.
\newblock Smpl: A skinned multi-person linear model.
\newblock In \emph{Seminal Graphics Papers: Pushing the Boundaries, Volume 2}, pages 851--866. 2023.

\bibitem[Lorensen and Cline(1998)]{lorensen1998marching}
William~E Lorensen and Harvey~E Cline.
\newblock Marching cubes: A high resolution 3d surface construction algorithm.
\newblock In \emph{Seminal graphics: pioneering efforts that shaped the field}, pages 347--353. 1998.

\bibitem[Mildenhall et~al.(2021)Mildenhall, Srinivasan, Tancik, Barron, Ramamoorthi, and Ng]{mildenhall2021nerf}
Ben Mildenhall, Pratul~P Srinivasan, Matthew Tancik, Jonathan~T Barron, Ravi Ramamoorthi, and Ren Ng.
\newblock Nerf: Representing scenes as neural radiance fields for view synthesis.
\newblock \emph{Communications of the ACM}, 65\penalty0 (1):\penalty0 99--106, 2021.

\bibitem[Mohammad~Khalid et~al.(2022)Mohammad~Khalid, Xie, Belilovsky, and Popa]{2022clipmesh}
Nasir Mohammad~Khalid, Tianhao Xie, Eugene Belilovsky, and Tiberiu Popa.
\newblock Clip-mesh: Generating textured meshes from text using pretrained image-text models.
\newblock In \emph{SIGGRAPH Asia 2022 conference papers}, pages 1--8, 2022.

\bibitem[Or-El et~al.(2022)Or-El, Luo, Shan, Shechtman, Park, and Kemelmacher-Shlizerman]{or2022stylesdf}
Roy Or-El, Xuan Luo, Mengyi Shan, Eli Shechtman, Jeong~Joon Park, and Ira Kemelmacher-Shlizerman.
\newblock Stylesdf: High-resolution 3d-consistent image and geometry generation.
\newblock In \emph{Proceedings of the IEEE/CVF Conference on Computer Vision and Pattern Recognition}, pages 13503--13513, 2022.

\bibitem[Pan et~al.(2023)Pan, Elgharib, Teotia, Tewari, Golyanik, Kortylewski, Theobalt, et~al.]{pan2023avatarstudio}
Mohit~Mendiratta Pan, Mohamed Elgharib, Kartik Teotia, Ayush Tewari, Vladislav Golyanik, Adam Kortylewski, Christian Theobalt, et~al.
\newblock Avatarstudio: Text-driven editing of 3d dynamic human head avatars.
\newblock \emph{arXiv preprint arXiv:2306.00547}, 2023.

\bibitem[Poole et~al.(2022)Poole, Jain, Barron, and Mildenhall]{poole2022dreamfusion}
Ben Poole, Ajay Jain, Jonathan~T Barron, and Ben Mildenhall.
\newblock Dreamfusion: Text-to-3d using 2d diffusion.
\newblock \emph{arXiv preprint arXiv:2209.14988}, 2022.

\bibitem[Radford et~al.(2021)Radford, Kim, Hallacy, Ramesh, Goh, Agarwal, Sastry, Askell, Mishkin, Clark, et~al.]{radford2021learning}
Alec Radford, Jong~Wook Kim, Chris Hallacy, Aditya Ramesh, Gabriel Goh, Sandhini Agarwal, Girish Sastry, Amanda Askell, Pamela Mishkin, Jack Clark, et~al.
\newblock Learning transferable visual models from natural language supervision.
\newblock In \emph{International conference on machine learning}, pages 8748--8763. PMLR, 2021.

\bibitem[Raj et~al.(2023)Raj, Kaza, Poole, Niemeyer, Ruiz, Mildenhall, Zada, Aberman, Rubinstein, Barron, et~al.]{dreambooth3d}
Amit Raj, Srinivas Kaza, Ben Poole, Michael Niemeyer, Nataniel Ruiz, Ben Mildenhall, Shiran Zada, Kfir Aberman, Michael Rubinstein, Jonathan Barron, et~al.
\newblock Dreambooth3d: Subject-driven text-to-3d generation.
\newblock \emph{arXiv preprint arXiv:2303.13508}, 2023.

\bibitem[Richardson et~al.(2023)Richardson, Metzer, Alaluf, Giryes, and Cohen-Or]{richardson2023texture}
Elad Richardson, Gal Metzer, Yuval Alaluf, Raja Giryes, and Daniel Cohen-Or.
\newblock Texture: Text-guided texturing of 3d shapes.
\newblock \emph{arXiv preprint arXiv:2302.01721}, 2023.

\bibitem[Rombach et~al.(2022)Rombach, Blattmann, Lorenz, Esser, and Ommer]{rombach2022high}
Robin Rombach, Andreas Blattmann, Dominik Lorenz, Patrick Esser, and Bj{\"o}rn Ommer.
\newblock High-resolution image synthesis with latent diffusion models.
\newblock In \emph{Proceedings of the IEEE/CVF conference on computer vision and pattern recognition}, pages 10684--10695, 2022.

\bibitem[Ruiz et~al.(2023)Ruiz, Li, Jampani, Pritch, Rubinstein, and Aberman]{ruiz2023dreambooth}
Nataniel Ruiz, Yuanzhen Li, Varun Jampani, Yael Pritch, Michael Rubinstein, and Kfir Aberman.
\newblock Dreambooth: Fine tuning text-to-image diffusion models for subject-driven generation.
\newblock In \emph{Proceedings of the IEEE/CVF Conference on Computer Vision and Pattern Recognition}, 2023.

\bibitem[Saharia et~al.(2022)Saharia, Chan, Saxena, Li, Whang, Denton, Ghasemipour, Ayan, Mahdavi, Lopes, Salimans, Ho, Fleet, and Norouzi]{Saharia2022PhotorealisticTD}
Chitwan Saharia, William Chan, Saurabh Saxena, Lala Li, Jay Whang, Emily~L. Denton, Seyed Kamyar~Seyed Ghasemipour, Burcu~Karagol Ayan, Seyedeh~Sara Mahdavi, Raphael~Gontijo Lopes, Tim Salimans, Jonathan Ho, David Fleet, and Mohammad Norouzi.
\newblock Photorealistic text-to-image diffusion models with deep language understanding.
\newblock 2022.

\bibitem[Sanghi et~al.(2022)Sanghi, Chu, Lambourne, Wang, Cheng, Fumero, and Malekshan]{sanghi2022clip}
Aditya Sanghi, Hang Chu, Joseph~G Lambourne, Ye Wang, Chin-Yi Cheng, Marco Fumero, and Kamal~Rahimi Malekshan.
\newblock Clip-forge: Towards zero-shot text-to-shape generation.
\newblock In \emph{Proceedings of the IEEE/CVF Conference on Computer Vision and Pattern Recognition}, pages 18603--18613, 2022.

\bibitem[Schwarz et~al.(2020)Schwarz, Liao, Niemeyer, and Geiger]{schwarz2020graf}
Katja Schwarz, Yiyi Liao, Michael Niemeyer, and Andreas Geiger.
\newblock Graf: Generative radiance fields for 3d-aware image synthesis.
\newblock \emph{Advances in Neural Information Processing Systems}, 33:\penalty0 20154--20166, 2020.

\bibitem[Shen et~al.(2021)Shen, Gao, Yin, Liu, and Fidler]{shen2021deep}
Tianchang Shen, Jun Gao, Kangxue Yin, Ming-Yu Liu, and Sanja Fidler.
\newblock Deep marching tetrahedra: a hybrid representation for high-resolution 3d shape synthesis.
\newblock \emph{Advances in Neural Information Processing Systems}, 34:\penalty0 6087--6101, 2021.

\bibitem[Song et~al.(2023)Song, Xu, Liu, Zhi, Shi, Zhang, Jiang, Feng, Sang, and Luo]{song2023agilegan3d}
Guoxian Song, Hongyi Xu, Jing Liu, Tiancheng Zhi, Yichun Shi, Jianfeng Zhang, Zihang Jiang, Jiashi Feng, Shen Sang, and Linjie Luo.
\newblock Agilegan3d: Few-shot 3d portrait stylization by augmented transfer learning.
\newblock \emph{arXiv preprint arXiv:2303.14297}, 2023.

\bibitem[Song et~al.(2022)Song, Han, Liu, Metaxas, and Elgammal]{song2022diffusion}
Kunpeng Song, Ligong Han, Bingchen Liu, Dimitris Metaxas, and Ahmed Elgammal.
\newblock Diffusion guided domain adaptation of image generators.
\newblock \emph{arXiv preprint arXiv:2212.04473}, 2022.

\bibitem[Tang et~al.(2023)Tang, Ren, Zhou, Liu, and Zeng]{tang2023dreamgaussian}
Jiaxiang Tang, Jiawei Ren, Hang Zhou, Ziwei Liu, and Gang Zeng.
\newblock Dreamgaussian: Generative gaussian splatting for efficient 3d content creation.
\newblock \emph{arXiv preprint arXiv:2309.16653}, 2023.

\bibitem[Wang et~al.(2023)Wang, Lu, Wang, Bao, Li, Su, and Zhu]{wang2023vsd}
Zhengyi Wang, Cheng Lu, Yikai Wang, Fan Bao, Chongxuan Li, Hang Su, and Jun Zhu.
\newblock Prolificdreamer: High-fidelity and diverse text-to-3d generation with variational score distillation.
\newblock \emph{arXiv preprint arXiv:2305.16213}, 2023.

\bibitem[Zhang et~al.(2023{\natexlab{a}})Zhang, Chen, Fu, Zhou, Yu, Wang, Fu, Chen, Lin, and Shen]{zhang2023styleavatar3d}
Chi Zhang, Yiwen Chen, Yijun Fu, Zhenglin Zhou, Gang Yu, Billzb Wang, Bin Fu, Tao Chen, Guosheng Lin, and Chunhua Shen.
\newblock Styleavatar3d: Leveraging image-text diffusion models for high-fidelity 3d avatar generation.
\newblock \emph{arXiv preprint arXiv:2305.19012}, 2023{\natexlab{a}}.

\bibitem[Zhang et~al.(2023{\natexlab{b}})Zhang, Chen, Yang, Qu, Wang, Chen, Long, Zhu, Du, and Zheng]{Avatarverse}
Huichao Zhang, Bowen Chen, Hao Yang, Liao Qu, Xu Wang, Li Chen, Chao Long, Feida Zhu, Kang Du, and Min Zheng.
\newblock Avatarverse: High-quality \& stable 3d avatar creation from text and pose.
\newblock \emph{arXiv preprint arXiv:2308.03610}, 2023{\natexlab{b}}.

\bibitem[Zhang et~al.(2023{\natexlab{c}})Zhang, Lan, Yang, Hong, Wang, Yeo, Liu, and Loy]{zhang2023deformtoon3d}
Junzhe Zhang, Yushi Lan, Shuai Yang, Fangzhou Hong, Quan Wang, Chai~Kiat Yeo, Ziwei Liu, and Chen~Change Loy.
\newblock Deformtoon3d: Deformable neural radiance fields for 3d toonification.
\newblock In \emph{Proceedings of the IEEE/CVF International Conference on Computer Vision}, pages 9144--9154, 2023{\natexlab{c}}.

\bibitem[Zhang et~al.(2023{\natexlab{d}})Zhang, Qiu, Lin, Zhang, Shi, Yang, Shi, Yang, Xu, and Yu]{dreamface}
Longwen Zhang, Qiwei Qiu, Hongyang Lin, Qixuan Zhang, Cheng Shi, Wei Yang, Ye Shi, Sibei Yang, Lan Xu, and Jingyi Yu.
\newblock Dreamface: Progressive generation of animatable 3d faces under text guidance.
\newblock \emph{arXiv preprint arXiv:2304.03117}, 2023{\natexlab{d}}.

\bibitem[Zhang et~al.(2023{\natexlab{e}})Zhang, Rao, and Agrawala]{zhang2023adding}
Lvmin Zhang, Anyi Rao, and Maneesh Agrawala.
\newblock Adding conditional control to text-to-image diffusion models, 2023{\natexlab{e}}.

\bibitem[Zhou et~al.(2021)Zhou, Xie, Ni, and Tian]{zhou2021cips}
Peng Zhou, Lingxi Xie, Bingbing Ni, and Qi Tian.
\newblock Cips-3d: A 3d-aware generator of gans based on conditionally-independent pixel synthesis.
\newblock \emph{arXiv preprint arXiv:2110.09788}, 2021.

\end{thebibliography}
}


\end{document}



\title{Supplementary Material for Paper:\\ DiffusionGAN3D: Boosting Text-guided 3D Generation and Domain Adaptation by Combining 3D GANs and Diffusion Priors}

\author{Biwen Lei, Kai Yu, Mengyang Feng, Miaomiao Cui, Xuansong Xie\\
Alibaba Group\\
{\tt\small \{biwen.lbw, jinmao.yk, mengyang.fmy, miaomiao.cmm\}@alibaba-inc.com, }\\
{\tt\small xingtong.xxs@taobao.com}
}
\maketitle




In this document, we present full-text prompts used in the experiments in Sec.~\ref{sec:prompt}, implementation details in Sec.~\ref{sec:implementation_details}, more visualization results in Sec.~\ref{sec:more_visulization}, the application of DiffusionGAN3D on real images (3D-aware editing and stylization) in Sec.~\ref{sec:applications}, discussions about the limitations of the proposed method and future work in Sec.~\ref{sec:limitation}.

\section{Full-Text Prompts} \label{sec:prompt}
In our paper, a shared additional positive prompt and the same negative prompt were applied in all the experiments including domain adaptation and text-to-avatar. The former was added after each primary prompt, and the latter served as a complete negative prompt for all experiments. Next, we provide the two shared prompts and all the primary prompts used in the experiments mentioned in the main paper and this document:
\begin{itemize}
\item additional positive prompt: “, sharp, 8K, skin detail, best quality, realistic lighting, good-looking, uniform light, extremely detailed"
\item negative prompt: “blurry, inaccurate identity, disproportionate, wrong anatomy, blurry face, ugly, bad face lighting"
\\
\\
\noindent\textbf{Main paper}
\item Figure 1.(Pixar): “Pixar style, a closeup of a person, cute, big eyes, Disney"
\item Figure 1.(Greek statue): “a closeup of a Greek statue, head"
\item Figure 1.(Joker): “joker, portrait" (“Zombie" in the 4th column of Figure 1. should be “Joker". It was a mistake.)
\item Figure 1.(Pixar(cat)): “Pixar style, a closeup of a cat, Disney"
\item Figure 1.(Fox, Pixar style): “Pixar style, a closeup of a fox, Disney"
\item Figure 1.(Golden statue): “Golden statue, a close-up of a cat"
\item Figure 1.(Hulk): “the Hulk, portrait"
\item Figure 1.(Batman): “the Batman, portrait"
\item Figure 1.(Obama): “Barack Obama, portrait"
\item Figure 6.(Pixar): “Pixar style, a closeup of a person, Disney"
\item Figure 6.(Lego): “Lego head"
\item Figure 8.: “Pixar style, a cute girl with black hair"
\item Figure 9.: “Pixar style, a closeup of a cat, Disney"
\item Figure 10.: “a close-up of a woman with green hair"
\item Figure 11.: “Link in Zelda, portrait"

\noindent\textbf{This document}
\item Fig.~\ref{fig:supp_domain_adaptation}.(Plaster statue): “a closeup of a plaster statue, head"
\item Fig.~\ref{fig:supp_domain_adaptation}.(Oil painting): “oil painting, a closeup of a person"
\item Fig.~\ref{fig:supp_domain_adaptation}.(Zombie): “zombie, head"
\item Fig.~\ref{fig:supp_domain_adaptation}.(Caricature): “caricature style, a closeup of a person"
\item Fig.~\ref{fig:supp_domain_adaptation}.(Cat, Pixar style): “Pixar style, a closeup of a cat, Disney"
\item Fig.~\ref{fig:supp_domain_adaptation}.(Fox, Pixar style): “Pixar style, a closeup of a fox, Disney"
\item Fig.~\ref{fig:supp_local_editing}.(Purple hair): “a closeup of a person with purple hair"
\item Fig.~\ref{fig:supp_local_editing}.(Blue eyes): “a closeup of a person with blue eyes"
\item Fig.~\ref{fig:supp_local_editing}.(Red lipstick): “Red lipstick, a closeup of a person"
\item Fig.~\ref{fig: supp_geometry}.: “Catwoman"
\end{itemize}

\section{Implementation Details} \label{sec:implementation_details}

\subsection{Architecture}

Our model and baselines are implemented using Pytorch 1.13.1 and Python 3.8. We utilize EG3D-based \cite{chan2022efficient} 3D GANs as our base generators, including PanoHead (head) \cite{an2023panohead}, EG3D-FFHQ $512\times512$ (face), EG3D-AFHQ 512x512 (cat) and AG3D (body) \cite{dong2023ag3d}. We use the open-source Diffusers \cite{von-platen-etal-2022-diffusers} library for the StableDiffusion \cite{rombach2022high} models. Training codes are built upon Stable-DreamFusion \cite{stable-dreamfusion}. In the progressive texture refinement stage, we employ Pytorch3D \cite{ravi2020pytorch3d} as the differentiable rendering framework.

\subsection{Training}

The latent mapping network and the decoder are frozen during optimization. We only optimize the weights of the triplane generator for both domain adaptation and text-to-avatar generation tasks. We train the model using the Adam optimizer and a learning rate of $1\times10^{-4}$. All the models (both domain adaptation and text-to-avatar) are trained for 10,000 steps with a batch size of 1. Training takes about 50 minutes on an A100 GPU with a memory cost of 14G. Note that, in the domain adaptation task, we calculate the relative distance loss between the triplane results of the current batch and the previous batch. In SDS, the denoising timestep t is uniformly sampled from a range $T_{SDS}=(T_{min}, T_{max})$, where $T_{min}=300$, $T_{max}=800$ as default in our experiments. And the CFG \cite{ho2022classifier} weight is set to 50 as default.

\subsection{Progressive Texture Refinement}

For 3D avatar generation task, we implement the texture refinement using uniformly selected $2k + 2$ ($2k$ indicates the views of the left and right sides, $2$ denotes the front view and the back view) azimuths and $j$ elevations. Specifically, we set $k=1, j=1$ for head generation, and $k=2, j=3$ for full-body generation. For the domain adaptation task, we only use 3 azimuths (-20, 0, 20 degrees) and a single elevation (0 degrees). The DDIM is adopted as the sampling method for the diffusion models and the number of denoising steps is set to 50. The denoising strengths of the image-to-image and inpainting are 0.6 and 0.4 respectively. The control weights for canny and depth conditions are both 1.

\section{More Visualization Results} \label{sec:more_visulization}

\subsection{Domain adaptation}

\begin{figure*}[t]
   \setlength{\belowcaptionskip}{-0.3cm}
   \setlength{\abovecaptionskip}{0.1cm}
\centerline{\includegraphics[scale=0.44]{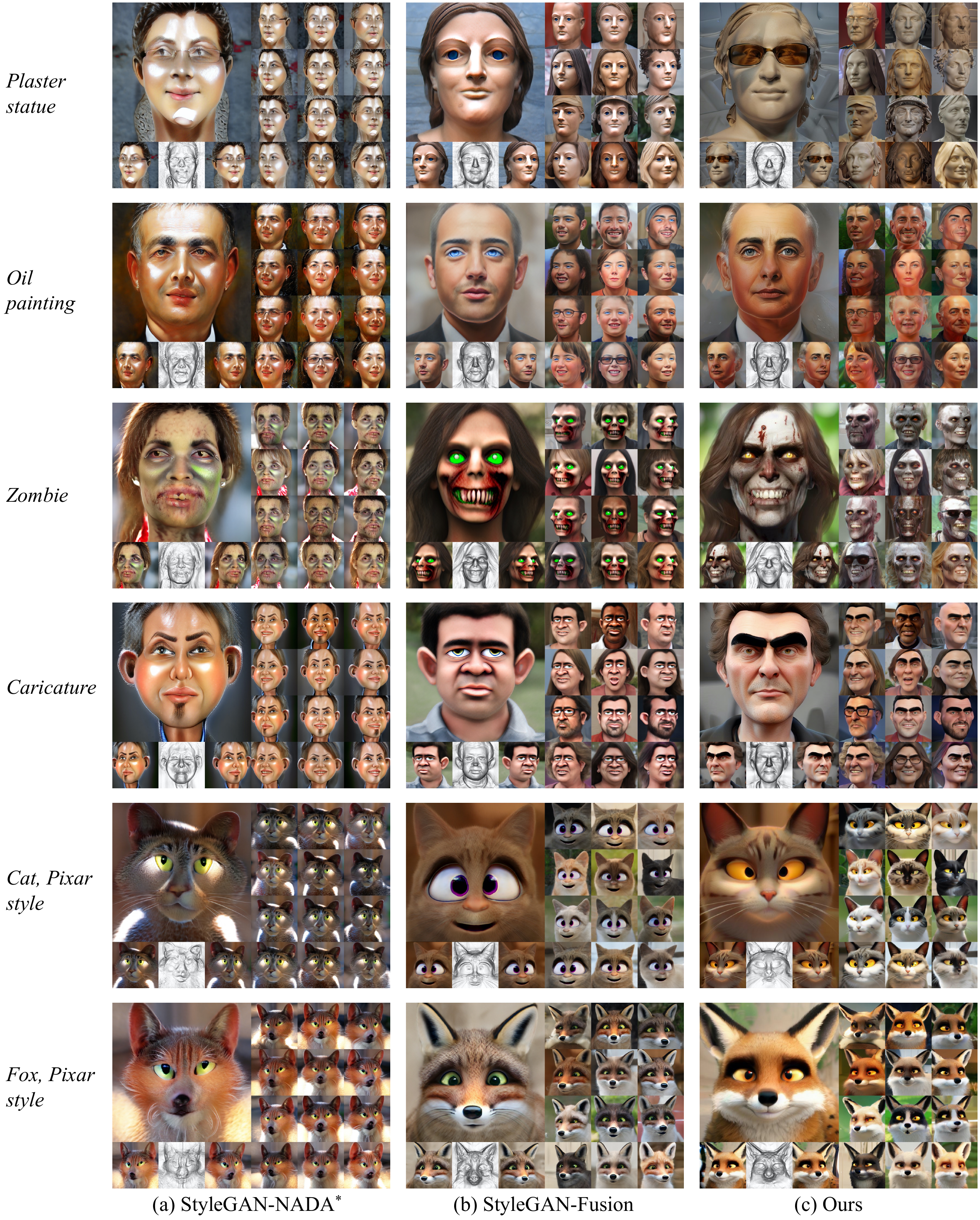}}
\caption{The qualitative comparisons for 3D domain adaptation on EG3D-FFHQ \cite{chan2022efficient} (the former four rows) and EG3D-AFHQ (the last two rows). (Zoom in for a better view.)}
\label{fig:supp_domain_adaptation}
\end{figure*}

More comparison examples of 3D domain adaptation on EG3D-FFHQ(face) and EG3D-AFHQ(cat) are shown in Fig.~\ref{fig:supp_domain_adaptation}. It can be seen that our method performs favorably against the others in terms of diversity, text-image correspondence, and texture quality. Please zoom in for a better view.

\subsection{Local Editing}

\begin{figure*}[t]
   \setlength{\belowcaptionskip}{-0.3cm}
   \setlength{\abovecaptionskip}{0.1cm}
\centerline{\includegraphics[scale=0.6]{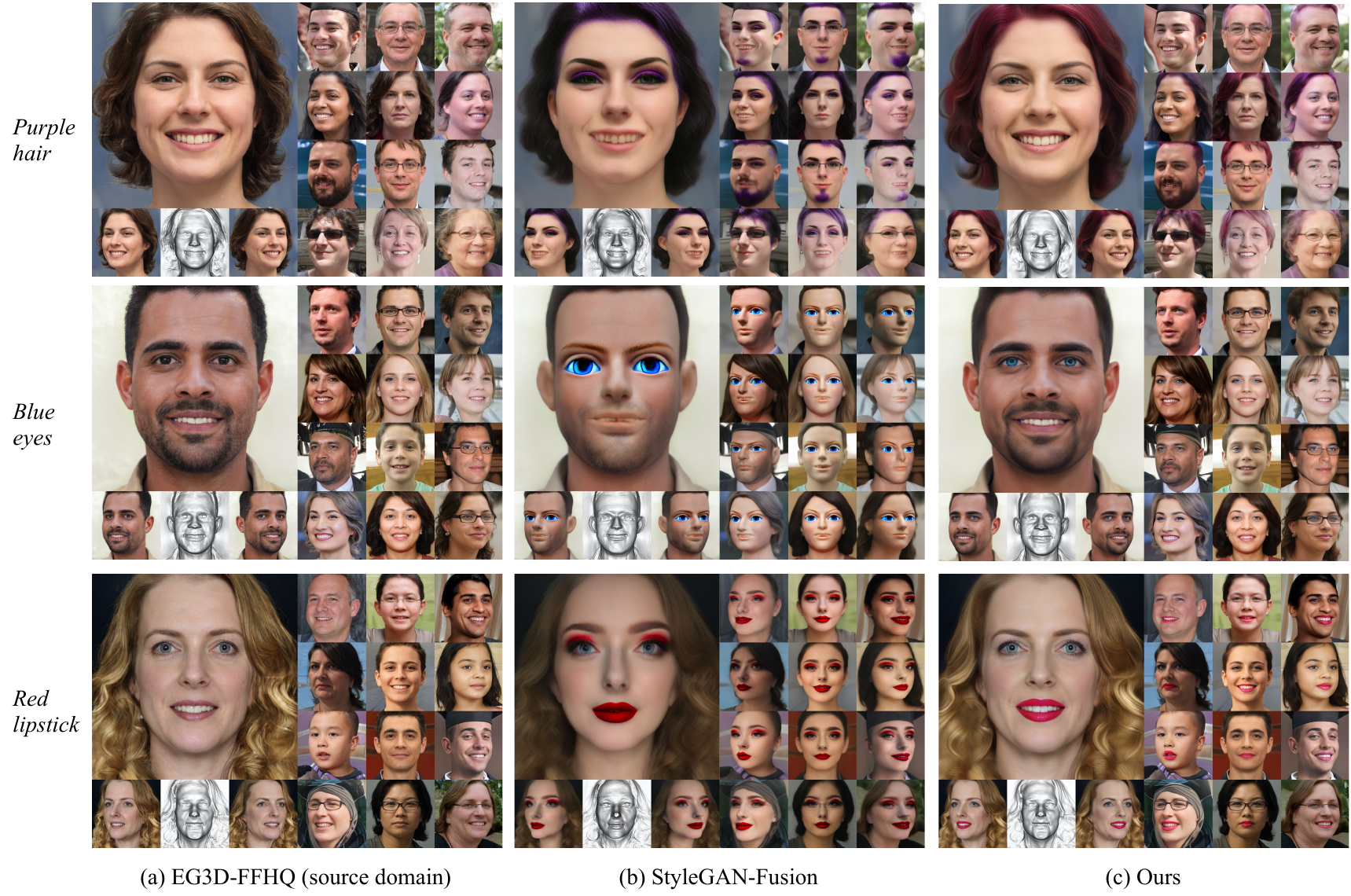}}
\caption{The qualitative comparisons for 3D-aware local editing on EG3D-FFHQ \cite{chan2022efficient}. (Zoom in for a better view.)}
\label{fig:supp_local_editing}
\end{figure*}

We compare the performance of our method and StyleGANFusion \cite{song2022diffusion} on some local editing scenarios. As shown in Fig.~\ref{fig:supp_local_editing}, StyleGAN-Fusion fails to preserve the details of the non-target region and leads to a global transformation. In this comparison, owing to the proposed diffusion-guided reconstruction loss, our method manages to precisely manipulate the target region while preserving the details of other regions and the overall identities.

\subsection{Text-to-Avatar}

\begin{figure*}[t]
   \setlength{\belowcaptionskip}{-0.3cm}
   \setlength{\abovecaptionskip}{0.1cm}
\centerline{\includegraphics[scale=0.64]{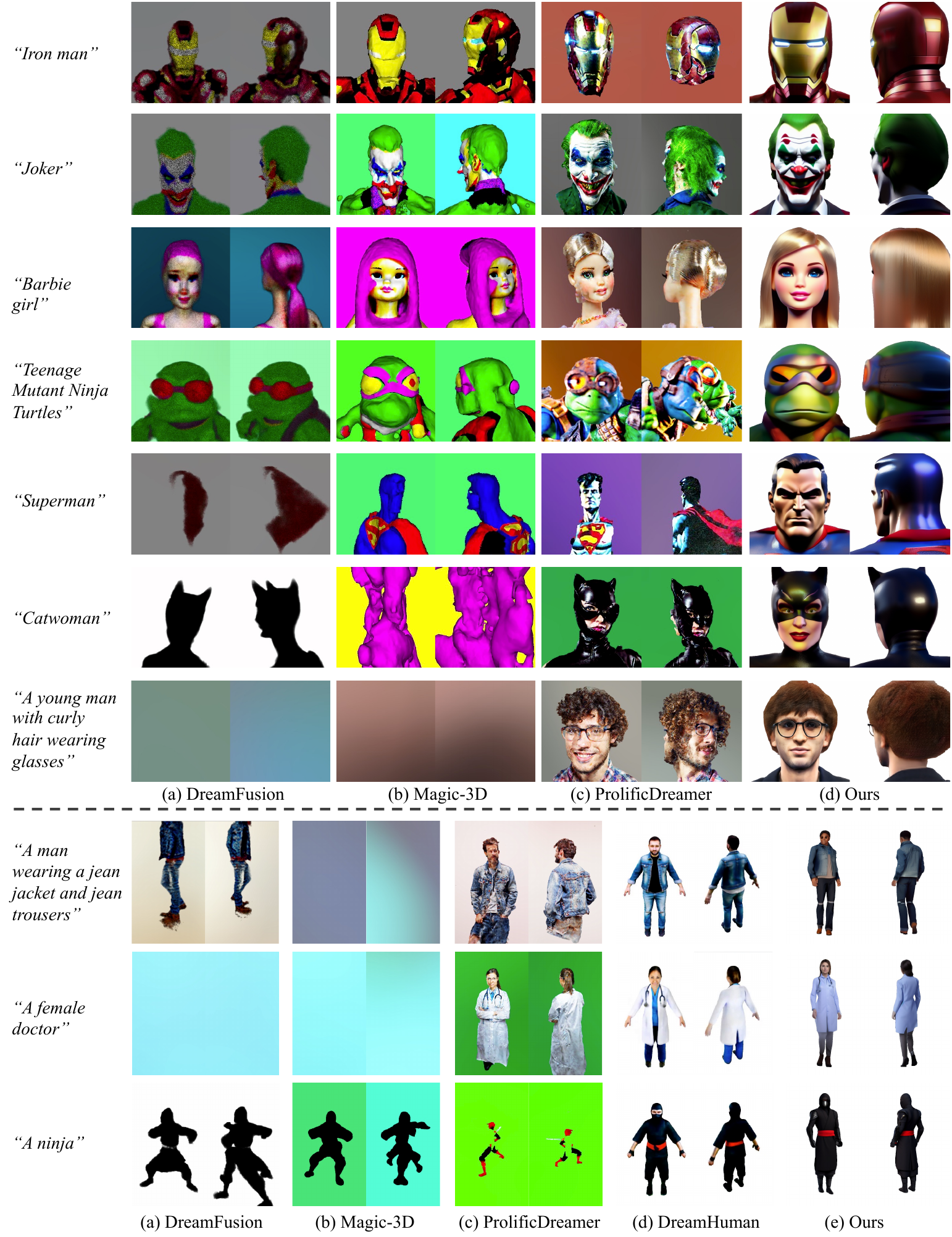}}
\caption{Visual comparisons on text-to-avatar task. The upper parts are the results of “head" and the lower parts are the results of “body". The empty images indicate convergence failure. (Zoom in for a better view.)}
\label{fig:supp_text-to-avatar}
\end{figure*}

We present more comparisons with other baselines on text-to-avatar tasks. As shown in Fig.~\ref{fig:supp_text-to-avatar}, although based on the 3D generators trained on realistic human images, the proposed method is capable of generating avatars across large domain gaps, showing great generation capability and stability. By contrast, the text-to-3D methods suffer from convergence failure, over-saturation, and incorrect geometry. DreamHuman \cite{dreamhuman} also has the problem of over-smoothed texture.

\begin{figure*}[t]
  \centering
  \resizebox{0.78\linewidth}{!}{
   \includegraphics{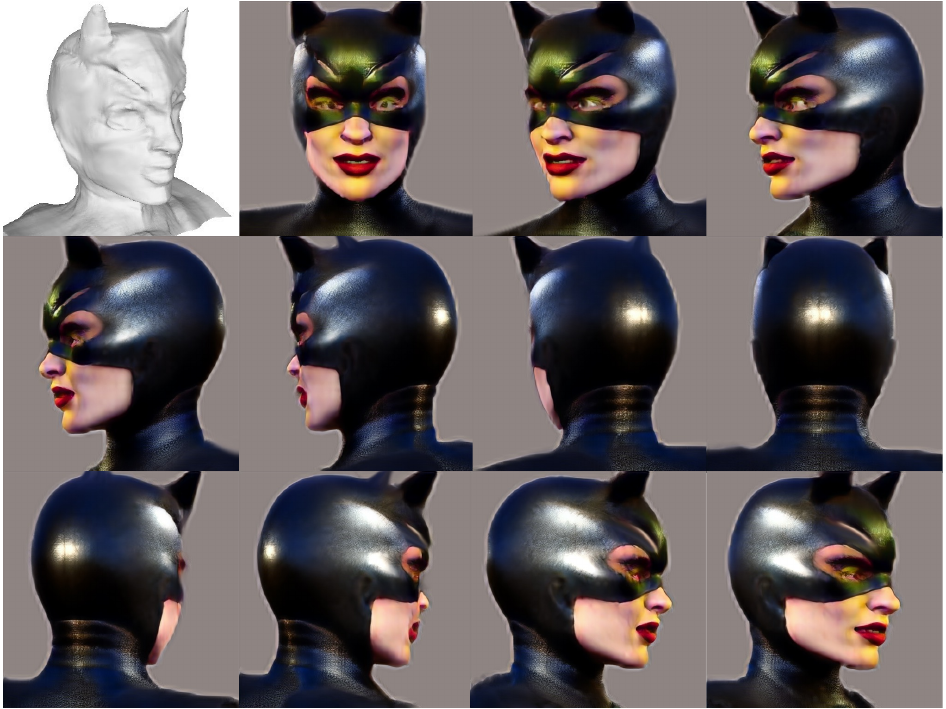} }
  \caption{The geometry and multi-view rendering results of the first stage of our method given the text “catwoman".}
  \label{fig: supp_geometry}
\end{figure*}

Besides, we give an example of the results of the first stage of our method without performing texture refinement. As shown in Fig.~\ref{fig: supp_geometry}, the generated geometry is smooth and the rendering results show decent quality and great view consistency.

\section{Applications} \label{sec:applications}

\begin{figure*}[t]
  \centering
  \resizebox{0.78\linewidth}{!}{
   \includegraphics{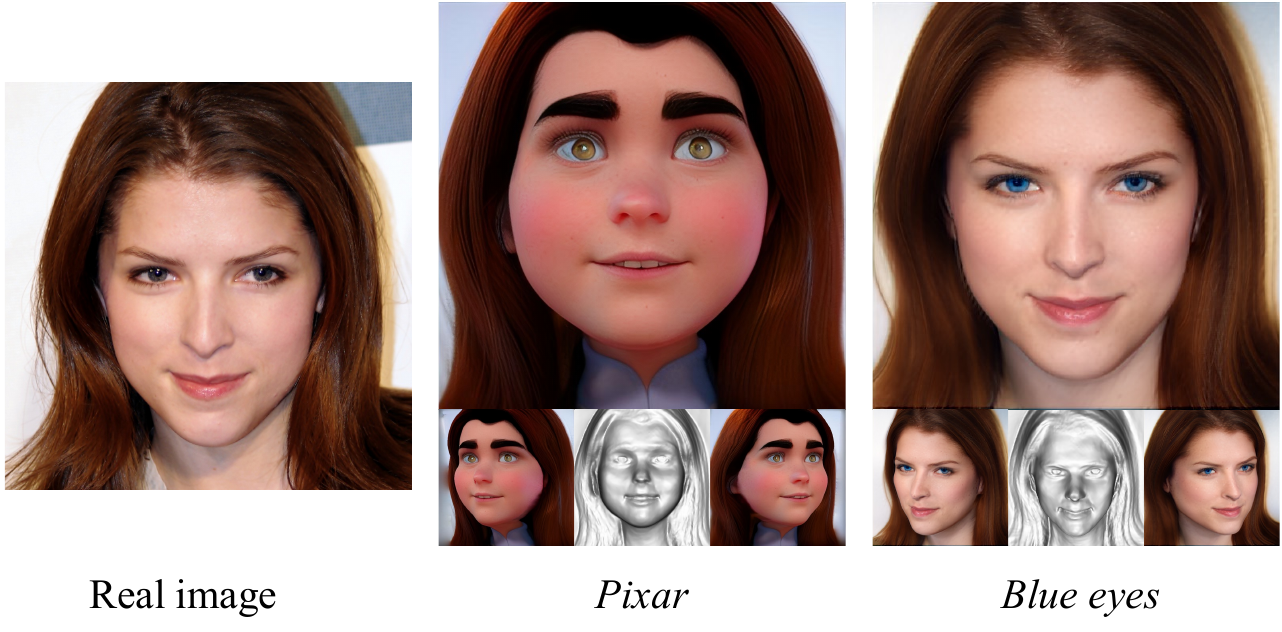} }
  \caption{An example of the application of our method on 3D-aware stylization and local editing.}
  \label{fig: supp_inversion}
\end{figure*}

Combined with the GAN inversion methods, DiffusionGAN3D is able to be applied to real face images to achieve text-guided 3D-aware stylization and local editing. We follow EG3D \cite{chan2022efficient} and employ PTI \cite{roich2021pivotal} to project the real image into the latent space. Specifically, the PTI includes two steps: the latent code inversion and generator finetuning. The former step generates the inverted latent code $w$ in $W+$ space and achieves rough reconstruction. The latter optimizes the generator and gets the finetuned generator $G'$ for accurate recovering. In our framework, for the stylization task with large domain gaps, such as Pixar and Lego style, we directly input $w$ into the stylized 3D GAN (such as the trained models in Fig.~\ref{fig:supp_domain_adaptation}) to achieve 3D stylization. For the stylization tasks with a small domain gap or the local editing tasks, we need to re-implement the domain adaptation on $G'$ to maintain the details and identity of the input image. Fig.~\ref{fig: supp_inversion} gives an example.

\section{Limitations and Future Work} \label{sec:limitation}

\noindent {\bf Limitations.} We summarize two limitations of our method. On one hand, the performance of DiffusionGAN3D relies on the base 3D generator. We conducted extensive experiments on EG3D-FFHQ(face), PanoHead(head), and AG3D(body) for both domain adaptation and text-to-avatar tasks. The results show that the models trained on EG3D-FFHQ and PanoHead exhibit superior performance than the model trained on AG3D. A possible solution is to boost the 3D generator itself with a high-quality dataset generated by diffusion models. On the other hand, the proposed methods cannot well handle the local editing with deformation, such as the prompt “fat face". We assume that the diffusion model itself cannot deal with local deformation well, and therefore the SDS \cite{poole2022dreamfusion} cannot provide clear guidance for 3D generators. 

\noindent {\bf Future Work.} Beyond addressing the limitations discussed above, we will further extend our method to achieve accurate, high-fidelity and animatable avatar generation from images or text descriptions for future work, conquering some challenging problems (such as modeling complex geometry and appearance).

{
    \small
    \bibliographystyle{ieeenat_fullname}
    \bibliography{main}
}